%% file: acl_latex.tex
\pdfoutput=1

\documentclass[11pt]{article}

\usepackage{acl}

\usepackage{times}
\usepackage{latexsym}
\usepackage{makecell}
\usepackage{booktabs}
\usepackage{subfigure}
\usepackage{amsmath,amsfonts,bm}
\usepackage{multicol}
\usepackage{multirow}
\usepackage{float}
\usepackage{enumitem}
\usepackage{tikz}
\usepackage{pgfplots}
\usepackage{xspace}
\usepackage{soul}

\usepackage[T1]{fontenc}

\usepackage[utf8]{inputenc}
\usepackage{xurl}

\usepackage{microtype}

\usepackage{inconsolata}

\title{EIT: Enhanced Interactive Transformer}

\author{
Tong Zheng$^{1}$\thanks{\xspace\xspace Equal Contribution.
},
  Bei Li$^{1*}$,
  Huiwen Bao$^{1,2*}$,
  \textbf{Tong Xiao$^{1,2}$\thanks{\xspace\xspace Corresponding author.}}
  \textbf{and Jingbo Zhu$^{1,2}$}\\
  $^1$School of Computer Science and Engineering, Northeastern University, Shenyang, China\\
  $^2$NiuTrans Research, Shenyang, China \\
  {\tt
        \{zhengtong12356, goodbaohuiwen\}@gmail.com, libei\_neu@outlook.com
  }\\
  {\tt
        \{xiaotong,zhujingbo\}@mail.neu.edu.cn
  }
}

\begin{document}
\maketitle
\begin{abstract}
Two principles: the \textit{complementary principle} and the \textit{consensus principle} are widely acknowledged in the literature of multi-view learning. However, the current design of multi-head self-attention, an instance of multi-view learning, prioritizes the complementarity while ignoring the consensus. To address this problem, we propose an enhanced multi-head self-attention (EMHA). First, to satisfy the \textit{complementary principle}, EMHA removes the one-to-one mapping constraint among queries and keys in multiple subspaces and allows each query to attend to multiple keys. On top of that, we develop a method to fully encourage consensus among heads by introducing two interaction models, namely inner-subspace interaction and cross-subspace interaction. Extensive experiments on a wide range of language tasks (e.g., machine translation, abstractive summarization and grammar correction, language modeling), show its superiority, with a very modest increase in model size. Our code would be available at: \url{https://github.com/zhengkid/EIT-Enhanced-Interactive-Transformer}. 
\end{abstract}

\section{Introduction}
    Transformer architectures \citep{Vaswani-etal-2017} have yielded promising results on a wide range of natural language processing tasks~\citep{devlin-etal-2019-bert,Brown-etal-2020-gpt3}. 
    A key factor contributing to their success is the multi-head self-attention network (MHSA), which enables efficient modeling of global dependencies among tokens in parallel. 
    Notably, instead of utilizing a single attention mechanism, MHSA uses an ensemble of attention models, each models a small subspace, and finally aggregates these results to the final one. The core idea is similar to subspace learning~\cite{Blum1998CoTraining} or multi-view learning~\cite{ Chaudhuri2009MultiViewclusteriung}.

    In the realm of multi-view learning, two fundamental principles guide the research: the \textit{complementary principle} and the \textit{consensus principle}~\cite{xu2013survey}. The \textit{complementary principle} emphasizes that each data view may possess unique knowledge not present in other views, prompting the use of multiple views for a comprehensive and accurate data description. On the contrary, the \textit{consensus principle} aims to maximize the agreement on multiple distinct views.  However, in the context of MHSA design, most studies predominantly focus on the \textit{complementary principle}. This oversight is evident in their encouragement of diverse information capture by different heads~\cite{li-etal-2018-multi-head, cui-etal-2019-mixed,zheng-etal-2024-partialformer} and the adoption of complex aggregation operations~\cite{li-etal-2019-information, Wang-etal-2020-Collaboration}. Some studies~\cite{MichelLN19-etal-16-better-than-8, clark-etal-2019-bert,voita-etal-2019-analyzing, behnke-heafield-2020-losing} even consider the high similarity among attention heads as a significant problem referred to as \textit{attention redundancy}.

    Although diversity is crucial in multi-view learning, \citet{dasgupta2001pac} has shown that simply fusing diverse outputs does not guarantee improved results: the probability of a disagreement of two independent hypotheses upper bounds the error rate of either hypothesis. The \textit{consensus principle} highlights the need to minimize disagreement among views to achieve better outcomes. In response to the \textit{consensus principle}, several studies~\cite{Kumar2011ACo-TrainingApproachforMulti-ViewSpectralClustering, Kumar2011Co-RegularizedMulti-ViewSpectralClustering} have focused on minimizing disagreement among views to achieve better outcomes. However, in the context of MHSA research, there is a tendency to prioritize complementarity over consensus among different attention heads. Here we ask a question: \textit{Can balancing these two principles benefit the design of MHSA mechanisms?}

    However, encouraging such a consensus in multi-head self-attention is challenging. In our preliminary experiments, we found that directly utilizing regularization terms can achieve this goal but cannot improve performance. Drawing inspirations from human behavior where group discussions and interactions foster consensus, we propose introducing interactions among different subspaces in MHSA to achieve consensus.  

    To this end, we propose a new multi-head self-attention variant: Enhanced Multi-Head Self-Attention, which encourages consensus among attention heads while guaranteeing to contain sufficient information. To ensure information sufficiency, we propose a novel many-to-many mapping scheme to generate numerous high-quality initial attention maps. This can generate more attention maps without suffering low-bottleneck problems~\cite{pmlr-v119-bhojanapalli20a}. On top of these sufficient attention maps, we propose two interaction components: \textit{inner-subspace interaction} (ISI) and \textit{cross-subspace interaction} (CSI). These hierarchical interaction modules fully encourage consensus among attention maps of different heads.

    The outcome of this work is an Enhanced Interactive Transformer~(\textsc{Eit}) architecture in that MHSA is replaced with Enhanced Multi-Head Attention (EMHA). Our proposed \textsc{Eit} has been demonstrated to be simple to implement and highly parameter efficient, yet it produces consistent performance improvements across a diverse set of tasks, including machine translation, grammar error correction, abstractive summarization, and language modeling. In addition, we have developed a computationally efficient variant of \textsc{Eit}, which, while still maintaining strong performance on several tasks, is better suited for low-latency industrial applications.

\section{Preliminary: Multi-Head Self-Attention}

\label{sec:Preliminary}

Multi-head self-attention (MHSA) is an efficient operation that can capture the interactions among tokens. Given an embedded input sequence $\mathbf{X} \in \mathbb{R}^{T \times d}$, MHSA is defined as follows:
\begin{gather}
\mathbf{A}^{i} = \mathrm{Softmax}(\frac{(\mathbf{X}\mathbf{W}_{Q}^{i})(\mathbf{X}\mathbf{W}_{K}^{i})^{\mathsf{T}}}{d^{k}})\label{eq:cal}\\
 \mathbf{O} = \sum_{i=1}^{M}\mathbf{A}^{i}\mathbf{X}\mathbf{W}_{V}^{i}\mathbf{W}_{O}^{i}
 \label{eq:aggrega}
\end{gather}
where $T$ denotes the sequence length, $d$ is the input embedding dimension, $d_k$ is the head dimension, $M$ is the number of head partition on representations, $\mathbf{W}_{Q}^{i}, \mathbf{W}_{K}^{i}, \mathbf{W}_{O}^{i} \in \mathbb{R}^{d \times d_k}$, $\mathbf{W}_{O}^{i} \in \mathbb{R}^{d_k \times d}$. $\mathbf{A}^{i}$ represents the attention distribution of $i$-th head and $\mathbf{O}$ represents the output of multi-head self-attention mechanism.  Without the special declaration, we use $\mathbf{Q}^{i}, \mathbf{K}^{i}, \mathbf{V}^{i}$ to refer to $\mathbf{X}\mathbf{W}_{Q}^{i},\mathbf{X}\mathbf{W}_{K}^{i},\mathbf{X}\mathbf{W}_{V}^{i}$, respectively, which denotes the query, key and value in $i$-th head.

\section{Enhanced Interactive Transformer}
\label{subsec:eit}
We design a novel Enhanced Interactive Transformer (\textsc{Eit}) in which we replace the multi-head self-attention with an Enhanced Multi-Head Attention mechanism (EMHA) that encourages consensus among different attention heads. Our method mainly modified Eq. (\ref{eq:cal}) but otherwise follows the standard Transformer.

\subsection{Many-to-Many Mapping Scheme}
Intuitively, to achieve better consensus, multi-head self-attention should first contain as much information as possible. To achieve this goal, a natural idea is to employ more attention heads in multi-head self-attention. However, multi-head self-attention with too many heads suffers from low bottleneck problems~\cite {pmlr-v119-bhojanapalli20a}, resulting in performance deterioration in practical applications. 

Although various strategies like \textit{attention expansion}~\cite{Shazeer-etal-2020, Zhou-etal-2021-refiner} have been proposed, the information captured in their attention maps remains limited due to an additional linear transformation step, which can introduce redundancy among the maps.

\input{Figure/many-to-many}
To alleviate this problem, we propose a novel many-to-many (M2M) mapping scheme that enables each query to attend to $M$ keys instead of a single key. As illustrated in Figure~\ref{fig:architecture}, four queries and four keys can serve as two components in a bipartite graph, and each element in a component, e.g., $\mathbf{Q}^1$, can interact with any elements in another component, e.g., $\mathbf{K}^1, \dots, \mathbf{K}^4$. Formally, supposing one with $M$ heads, the $i$-th attention map can be formally calculated as:
\begin{equation}
\begin{split}
     \mathbf{S}^{i} = \frac{\mathbf{Q}^{\lfloor (i-1)/M+1 \rfloor}({\mathbf{K}^{(i-1)\%M+1}})^{\mathsf{T}}}{\sqrt{d_k}} 
\label{eq:5}
\end{split}
\end{equation}
\noindent where $i \in \{1,\dots,M^2\}$, $\mathbf{S}^i \in \mathbb{R}^{T\times T}$ is the attention maps without softmax, $\lfloor \rfloor$ is the round down operation and \% is the mod operation. For example, $\mathbf{S}^4$ is computed by $\mathbf{Q}^1$ and $\mathbf{K}^4$ when $M=4$.

\paragraph{Discussion.} M2M demonstrates an increased capacity to generate M times the number of attention maps when given identical input. This enhanced capability can be attributed to the effective utilization of a many-to-many mapping strategy by M2M, which fully leverages the original head features, such as $\mathbf{Q}$ and $\mathbf{K}$. Notably, this approach successfully avoids the production of similar attention maps by employing a dot-multiplication strategy to directly generate the attention maps~(See Figure \ref{fig:dynamics_similarity}). By avoiding the generation of redundant attention maps, M2M improves its ability to capture diverse and distinct patterns in the input data. As a result, it facilitates the subsequent creation of more comprehensive and informative representations. This module can also be viewed as a strategy to enhance \textit{complementary principle}.

\subsection{Dual Enhanced Interaction}
As aforementioned, M2M enlarges the information capacity, which provides a prerequisite for encouraging consensus among different heads. To encourage consensus, a simple idea is to directly add a linear transformation among attention maps~\cite{Shazeer-etal-2020, Zhou-etal-2021-refiner, Wang-etal-2020-Collaboration}. While these methods can achieve performance improvements in vanilla Transformer settings, they are unsuitable in our framework.  
One key factor is that our framework contains a substantial amount of information; however, it also incorporates certain elements of noise. Such a coarse interaction fails to attain a satisfactory consensus.

To address this problem, we propose a more refined solution that is able to differentiate between relevant and irrelevant information, discarding the latter while fully utilizing the former. Two kinds of interactions among those attention maps are introduced hierarchically, the inner-subspace interaction and cross-subspace interaction. 

\input{Figure/dual_enhanced_interaction}

\paragraph{Two Relationships.} We begin with identifying two important relationships: inner-subspace interaction (ISI) relationship and cross-subspace interaction (CSI) relationship. As illustrated in Figure \ref{fig:architecture}, the inner-subspace interaction (ISI) relationship describes the connection among the attention maps generated by the same query, e.g., attention maps in the block of the same color. These attention maps own a closer relationship. The cross-subspace interaction (CSI) relationship describes the collaboration among different heads, which exists in the attention maps generated by different queries, e.g., attention maps from blocks of different colors. 

\paragraph{Inner-Subspace Interaction Modeling.}
One can adopt the standard convolution operation via batch transformation. However, such a way 
ignores the difference among the ISI relationship constrained by different queries, e.g., the ISI relationship in red block and blue block in Figure \ref{fig:architecture}. It is desirable to preserve and enhance this distinction. To more efficiently model the interaction within subspaces, we therefore adopt group convolutions~\citep{Krizhevsky-etal-2012-alexnet}, which use separate parameters to process features from different groups.

Suppose $\bf f(\cdot)$ as a single layer group convolution. As illustrated in Figure \ref{fig:dei}, given the output of M2M, namely $\mathbf{S}$, as input, ISI sub-module computed as:
    \begin{equation}
    {{\dot{\bf{S}}}} = \bf f^{(1)}(\bf{ReLU}(f^{(0)}(  {{{\bf{S}}} } )))\label{eq.isi}
    \end{equation}
where ${{\dot{\bf{S}}}} \in \mathbb{R}^{M \times T \times T}$ is the output of the ISI sub-module. We use $M^{H_{isi}}$ to represent the intermediate number of heads in ISI sub-module and set the number of groups in group convolutions to $M$. 

Finally, we can obtain $M$ high-quality attention maps that effectively retain the benefits of using a larger number of attention heads while discarding irrelevant information. Such a process is another key for Transformer to benefit from more heads and is unique to our work.  

\paragraph{Cross-Subspace Interaction Modeling.}
To efficiently model the cross-subspace interaction, we adopt two-layer convolutions accompanied by the ReLU activation to consist this sub-module.

Let us denote $\bf g(\cdot)$ as a single layer convolution. As illustrated in Figure \ref{fig:dei}, given the output of ISI, namely $\dot{\mathbf{S}}$, as input, CSI sub-module computed as: \begin{equation}
    {{\ddot{\bf{S}}}} = \bf g^{(1)}(\bf{ReLU}(g^{(0)}(  {{\dot{\bf{S}}} }))) \label{eq.csi}
\end{equation}
where ${{\ddot{\bf{S}}}} \in \mathbb{R}^{M \times T \times T}$ is the output of the CSI sub-module. We use $M^{H_{csi}}$ to represent the intermediate number of heads in CSI sub-module. Finally, we can obtain $M$ final attention maps that fully leverage the benefits of each head.

\subsection{Efficient Version of \textsc{Eit}}
\label{subsec:eeit}
\input{Figure/eeit}

Despite the theoretical computational efficiency and parametric efficiency of group convolutions, they slow down the training in practice \citep{Ma-etal-2018}. To alleviate this issue, we provide another efficient version of \textsc{Eit}, namely \textsc{E-Eit}, by simplifying the design of dual enhanced interaction. As illustrated in Figure~\ref{fig:eeit}, both ISI and CSI adopt a single-layer operation. Formally, the dual enhanced interactions are computed as: 
\begin{equation}
    { {\ddot{\bf{S}}}} = \bf g^{(0)}(\bf{ReLU}(f^{(0)}\left(  {{{\bf{S}}} } \right))) ,\label{eq.eeit}
\end{equation}
where $\bf{ReLU}(f^{(0)}\left(  {{{\bf{S}}} } \right))$, namely as  ${\dot{\bf{S}}}$, $\in \mathbb{R}^{ M^{H} \times T \times T}$ and ${\ddot{\bf{S}}} \in \mathbb{R}^{ M \times T \times T}$  and $M^{H}$ is a hyper-parameter, e.g. we set it as 32 for the base configuration.
In this way, \textsc{E-Eit} avoids parts of memory consumption and somehow improves the computational efficiency. 

\section{Experiment Settings}
We evaluated our \textsc{Eit} on four widely used benchmarks\footnote{We also evaluated our \textsc{Eit} variants on some task beyond natural language processing in Appendix.}: 1) Machine Translation, 2) Grammar Error Correction, 3) Abstractive Summarization, and 4) Language Modeling. The detailed architecture setups, training setups and evaluation setups were presented in Appendix \ref{Appendix:Experimental_Setups}.

\subsection{Machine Translation}
\label{sec:settings_mt}
\paragraph{Dataset.} We selected four widely used corpus: WMT'14 English-German (En-De) translations\footnote{\url{https://github.com/facebookresearch/fairseq/tree/main/examples/scaling_nmt.}} (a large-scale dataset, 4.5M training sentence pairs), WMT'16 English-Romanian (En-Ro) translations (a small-scale dataset, 610K training sentence pairs), WMT'17 English-German (En-De) translations and WMT'17 German-English (De-En) translations\footnote{\url{https://data.statmt.org/wmt17/translation-task/preprocessed/}}. The validation and test sets are \textit{newstest2013} and \textit{newstest2014}, respectively. For the En-Ro task, it consists of 610K training sentence pairs. We preprocessed the data as the setups in \citet{Mehta-2021-delight}\footnote{\url{https://github.com/sacmehta/delight/blob/master/readme_files/nmt/wmt16_en2ro.md}}. We performed shared BPE operations on both datasets to overcome the out-of-vocabulary (OOV) problem. Concretely, we set the size of BPE operations to 32K and 20K for En-De and En-Ro datasets, resulting in a shared vocabulary with sizes of 34040 and 19064, respectively.

\paragraph{Models. } Our model architectures were based on Transformer~\citep{Vaswani-etal-2017}. We provided three configurations, namely \textit{base}, \textit{big} in \citet{Vaswani-etal-2017}, and \textit{deep} in \citet{li-etal-2020-shallow,li2021gpkd}.  We adopted a pre-normalization strategy~\citep{wang-etal-2019-learning-deep} considering training stability under different configurations. 

\paragraph{Training \& Evaluation.}
We implemented our models using Fairseq~\citep{ott-etal-2019-fairseq}. Training employed 8 GEFORCE RTX 3090 cards for WMT'14 En-De and 4 cards for WMT'16 En-Ro, with batch sizes of 65536 and 16384, respectively. In the En-De task, we completed 50K updates for \textit{base}, 50K for \textit{deep}, and 100K for \textit{big} models. We utilized Adam~\citep{Kingma-etal-2015} with $\mathrm{adam}_{\beta}$ (0.9, 0.997) as the optimizer, an \textit{invert sqrt} learning rate scheduler with a rate of 0.002 and 16000 warmup updates, and 0.1 label smoothing for all experiments. During evaluation, we used 4 beams with a length penalty of 0.6 for En-De and 5 beams with a length penalty of 1.3 for En-Ro. For evaluation metrics, we utilized \textit{multi-BLEU}\footnote{\url{
https://github.com/moses-smt/mosesdecoder/blob/maste
r/scripts/generic/multi-bleu.perl}}~\cite{papineni-etal-2002-bleu}, COMET-22\footnote{\url{https://github.com/Unbabel/COMET}}~\cite{rei-etal-2022-comet} and sacreBLEU~\cite{post-2018-call} scores. We ran each experiment three times and reported the average score.

\begin{table*}[t!]
\centering
\renewcommand{\arraystretch}{1}
\small
\setlength{\tabcolsep}{2pt}
\resizebox{\linewidth}{!}{\begin{tabular}{clrrccccrcc}
\toprule
\multirow{2.5}{*}{\textbf{Type}} & \multirow{2.5}{*}{\textbf{Model}} & \multicolumn{5}{c}{\textbf{WMT'14 En-De}} & \multicolumn{4}{c}{\textbf{WMT'16 En-Ro}} \\
\cmidrule(r){3-7} \cmidrule(r){8-11}
& &Depth & $\mathbf{\theta}$ (M) & BLEU & sBLEU & COMET-22 &Depth & $\mathbf{\theta}$ (M) & BLEU & COMET-22 \\
\midrule
\multirow{7}{*}{\makecell[c]{Multi-Head}}
& Refiner~\cite{Zhou-etal-2021-refiner} & 6-6& - & 27.62 & - & - & 6-6 & 54 & 34.25 & - \\
& Talking-Head~\cite{Shazeer-etal-2020} & 6-6& - & 27.51 & - & - & 6-6 & 54 & 34.35 & - \\
& Collaboration~\cite{Wang-etal-2020-Collaboration} & 6-6& - & 27.55 & - & - & 6-6 & 54 & 34.64 & - \\
& \textsc{DyRouting}~\cite{li-etal-2019-information} & 6-6 & 297 & 28.96 & - & - & - & - & - & - \\
& \textsc{Disagree}~\cite{li-etal-2018-multi-head} & 6-6 & - & 29.28 & - & - & - & - & - & - \\
& MoA~\cite{Zhang-etal-2022-moa} & 6-6& 200 & 29.40 & - & - & 6-6 & 56 & 34.39 & - \\
& FISHformer~\cite{nguyen2022improving} & 6-6 & - & 29.57 & - & - & 6-6 & 49 & 34.42 & - \\
\midrule
\multirow{3}{*}{\makecell[c]{Localness}}
& \textsc{Dman}~\cite{fan-etal-2021-mask} & 6-6 & 211 & 28.97 & 27.8 & - & -& - & 34.49 & - \\
& \textsc{Csan}~\cite{yang-etal-2019-convolutional} & - & - & 28.74 & - & - & - & - & - & - \\
& \textsc{Umst}~\cite{li-2022-umst} & 6-6 &  242 & 29.75 & - & - & 6-6 & 60 & 34.81 & - \\
\midrule
\multirow{3}{*}{\makecell[c]{Other Baselines}}
& Transformer in \citet{liu-etal-2020-multilingual-denoising} & - & - & - & - & - & -& - & 34.30 & - \\
& Transformer in \citet{kasai2020non} & - & - & - & - & - & - & - & 34.16 & - \\
& Delight~\cite{Mehta-2021-delight} & - &  - & - & - & - & - & 53 & 34.70 & - \\
\midrule
\multirow{9}{*}{\makecell[c]{Our System }}
& Transformer base & 6-6 & 61.56 & 27.13 & 26.0 & 82.23 & 6-6 & 53.90 & 34.23 & 81.39 \\
& \textsc{Eit} base & 6-6& 61.63 & \bf 28.00 & \bf 26.9 & \bf 83.06 & 6-6& 53.98 & \bf 35.10 & \bf 82.18 \\
& \textsc{E-Eit} base & 6-6& 61.57 & 27.72 & 26.7 & 82.77 & 6-6 & 53.92 & 35.01 & 82.05 \\
\cmidrule(r){2-11}
& Transformer deep & 48-6& 193.96 & 29.60 & 28.5 & 84.21 & 24-6 & 110.64 & 35.00 & 82.11 \\
& \textsc{Eit} deep & 48-6 & 194.32 & \bf 30.25 & \bf 29.2 & \bf 84.74 & 24-6 & 111.09 & \bf 35.40 & \bf 82.46 \\
& \textsc{E-Eit}  & 48-6& 194.14 & 30.16 & 29.1 & 84.67 & 24-6& 110.73 & 35.35 & 82.55 \\
\cmidrule(r){2-11}
& Transformer big & 6-6& 211.22 & 28.80 & 27.7 & 83.53 & 6-6& 195.88 & 34.44 & 81.63 \\
& \textsc{Eit} big & 6-6& 211.55 & \bf 29.79 & \bf 28.7 & \bf 84.36 & 6-6& 196.40 & \bf 34.91 & \bf 82.15 \\
& \textsc{E-Eit} big & 6-6& 211.30 & 29.61 & 28.5 & 84.24 & 6-6& 195.97 & 34.67 & 81.80 \\
\bottomrule
\end{tabular}}
\caption{\label{tab:en-de-en-ro}
Results on WMT'14 En-De and WMT'16 En-Ro Tasks.}
\end{table*}

\subsection{Grammar Error Correction}
\label{sec:result_gec}
\paragraph{Dataset.}
We also assessed \textsc{Eit}'s effectiveness for grammar error correction, a crucial natural language processing application. Our experiments were conducted on the CONLL dataset, comprising 827K training sentences. Following the setup in \citep{Chollampatt-etal-2018-aaai}, we incorporated the word-level dropout technique \citep{sennrich-etal-2016-edinburgh} to mitigate overfitting. We configured BPE operations to 30K.
\paragraph{Models.}
We selected the Transformer~\citep{Vaswani-etal-2017} and \textsc{Surface}~\citep{Liu-etal-2021-surface} for comparison. These architectures adhere to the Transformer-base configuration outlined in \citet{Vaswani-etal-2017}.

\paragraph{Training \& Evaluation.}
We trained grammar error correction models on 8 GEFORCE RTX 3090 cards, using a batch size of 65536 and completing 14K total updates. Further training specifics can be found in Table \ref{tab:training}. For testing, we configured the number of beams to 6 and the length penalty to 0.6.

\subsection{Abstractive Summarization}
\label{sec:result_as}
\paragraph{Dataset.} 
We also tested the effectiveness of \textsc{Eit} on abstractive summarization task, a task relying on the ability of modeling long dependency. Shared BPE operations of 30K were applied to the training data, resulting in a vocabulary of 32,584 words.

\paragraph{Models.}
Our models were all under base configuration, e.g., embedding dimension, hidden dimension, $M$ are set to 512, 2048 and 8, respectively. 

\paragraph{Training \& Evaluation.} We trained abstractive summarization models on 8 GEFORCE RTX 3090 cards, utilizing a batch size of 131,072 and completing 30,000 total updates, the same setting in \citet{li-etal-2022-ode}. We incorporated a weight decay strategy with a ratio of 0.0001. We set warming updates to 16000. For testing, we configured 4 beams and a length penalty of 2.0, with minimum and maximum lengths set to 55 and 140, respectively.

\subsection{Language Modeling}
\label{sec:result_lm}
\paragraph{Dataset.} We assessed \textsc{Eit} in a language modeling task using WikiText-103 to investigate its capacity for handling long dependencies. The training, validation, and test sets encompass 103 million words (from 28,000 articles), 218,000 words, and 246,000 words, respectively. We adhered to the official preprocessing procedure \citep{ott-etal-2019-fairseq}\footnote{\url{https://github.com/facebookresearch/fairseq/tree/main/examples/language_model}}.
\paragraph{Models.} We chose the Adaptive Input Transformer~\citep{baevski2018adaptive} as the baseline model. All models are 8-layer models with 8 heads.
\paragraph{Training \& Evaluation.} The training and evaluation settings adhered to the standard PyTorch~\citep{ott-etal-2019-fairseq} language modeling guidelines. We trained both the baseline and \textsc{Eit} with 286,000 updates. During evaluation, we selected the checkpoint with the best performance on the validation set. Parameters such as max-tokens, max-sentences, and context-window were set to 3072, 1, and 2560, respectively.

\section{Experiments Results}

\subsection{Machine Translation}
\label{sec:result_mt}
\paragraph{Results of WMT'14 and WMT'16.}Table \ref{tab:en-de-en-ro} displays the results on WMT'14 En-De and WMT'16 En-Ro tasks. First, we can see that our \textsc{Eit} variants demonstrate superior performance compared to the vanilla Transformer across various configurations on both tasks. This indicates the effectiveness of \textsc{Eit} variants. Notably, \textsc{E-Eit}, an alternative to satisfy the low latency of industrial applications, can deliver competitive results compared with the full version while maintaining fast processing speeds. 

Besides, Our \textsc{Eit} can beat all selected methods of head modification and localness modeling, including the latest methods such as MoA~\citep{Zhang-etal-2022-moa}, Fishformer~\citep{nguyen2022improving}, UMST~\citep{li-2022-umst}, on both datasets. 
This highlights the fact that focusing on a single aspect, such as complementarity, is inadequate for achieving optimal results. By contrast, considering both complementarity and consensus leads to better performance.

\paragraph{Results of WMT'17.} Table \ref{tab:result_wmt_17} displays the results on WMT'17 En-De and De-En tasks. We can make similar observations as with the WMT’14 and WMT’16 tasks. Specifically, \textsc{Eit} can achieve BLEU points of 29.58 and 35.62 on WMT'17 En-De and De-En tasks, respectively, outperforming vanilla Transformer by BLEU points of 0.99 and BLEU points of 0.58. COMET-22 scores can further support this observation.

\begin{table}[t]
    \centering
    \setlength{\tabcolsep}{5pt}
    \resizebox{\linewidth}{!}{\begin{tabular}{@{}lcccc@{}}
        \toprule
        \multirow{2}{*}{\textbf{Method}} & \multicolumn{2}{c}{\textbf{WMT17 En-De}} & \multicolumn{2}{c}{\textbf{WMT17 De-En}} \\
        \cmidrule(lr){2-3} \cmidrule(lr){4-5}
        & BLEU & COMET-22 & BLEU & COMET-22\\
        \midrule
        Transformer & 28.59 & 81.11 & 35.04 & 82.48 \\
        \textsc{Eit} & 29.58  & 81.83  & 35.62  & 82.92  \\
        \bottomrule
    \end{tabular}}
    \caption{Results on WMT'17 En-De and De-En Tasks.}
    \label{tab:result_wmt_17}
\end{table}
\begin{table}[t!]
\renewcommand{\arraystretch}{1}
\centering
\small
\setlength{\tabcolsep}{1.5pt}
\resizebox{1\linewidth}{!}{
\begin{tabular}{lccc}
\toprule
\bf Model & \bf Precision &\bf Recall&  $\mathbf{F}_{0.5}$ \\
\midrule

Transformer~\ddag & 64.84 & \bf 36.61& 56.18\\
Talking-Head~\cite{Shazeer-etal-2020} & 64.32 &	36.07 &	55.61\\
\textsc{Surface}~\citep{Liu-etal-2021-surface} & 66.80  &35.00  & 56.60\\
\textsc{Eit}  & \bf 69.98& 32.80& 57.05\\
\textsc{E-Eit}  & 69.85 & 33.36& \bf 57.31 \\
\bottomrule
\end{tabular}
}
\caption{Results on the correction task.}
\label{tab:correction}
\end{table}
\begin{table}[t!]
\renewcommand{\arraystretch}{1}
\centering
\small
\setlength{\tabcolsep}{2pt}
\resizebox{1\linewidth}{!}{
\begin{tabular}{lrrr}
\toprule
\textbf{Model} & \textbf{RG-1}& \textbf{RG-2}& \textbf{RG-L} \\
\midrule

Transformer~\ddag &  40.84  & 18.00  & 37.58 \\
Talking-Head~\cite{Shazeer-etal-2020} & 41.26	& 18.34	& 38.06\\
 \midrule
PG-Net~\citep{see-etal-2017-get} & 39.53 & 17.28& 36.38 \\
\textsc{Mady}~\citep{Wang-etal-2021-mady} & 40.72& 17.90 &37.21  \\
\textsc{Dman}~\citep{fan-etal-2021-mask} & 40.98& 18.29&37.88  \\
\textsc{Bottom-up}~\citep{gehrmann-etal-2018-bottom} & 41.22& 18.68 & \bf 38.34\\
\textsc{Surface}~\citep{Liu-etal-2021-surface} &  41.00 & 18.30 & 37.90 \\
\textsc{Eit}  & \bf 41.62 & \bf 18.70 & 38.33 \\
\textsc{E-Eit}  & 41.58 & 18.63 & 38.28 \\
\bottomrule
\end{tabular}
}
\caption{Results on the summarization task.}
\label{tab:abstract}
\end{table}

\subsection{Grammar Error Correction}
\label{sec:result_gec}

Table \ref{tab:correction} presents the results of the CONLL dataset's test set. Both \textsc{Eit} and \textsc{E-Eit} outperform the standard Transformer, showing improvements of 0.87 and 1.13 in terms of $\mathrm{F}_{0.5}$, respectively. Compared to the strong baseline \textsc{Surface}, our methods (\textsc{Eit} and \textsc{E-Eit}) still outperform it by 0.45 and 0.71 $\mathrm{F}_{0.5}$ points, respectively. Importantly, both \textsc{Eit} and \textsc{E-Eit} require negligible extra parameters, less than 0.1M, indicating their enhanced expressive power. Notably, the Talking Heads model underperforms, possibly due to imperfect hyper-parameters, needing more fine-tuning.

An interesting observation is that \textsc{Eit} variants seem to trade recall for precision. This behavior is due to \textsc{Eit}'s foundation in both complementary and consensus principles, which naturally generate more precise attention maps by filtering out uncertain information. As a result, \textsc{Eit} primarily makes corrections where it is most confident.

\subsection{Abstractive Summarization}
\label{sec:result_as}

Table \ref{tab:abstract} shows results on the test set of CNN-DailyMail. We can see \textsc{Eit} can achieve scores of 41.62 ROUGE-1 points, 18.70 ROUGE-2 points and 38.33 ROUGE-L points, outperforming the standard Transformer by 0.78, 0.70 and 0.75  in terms of ROUGE-1, ROUGE-2 and ROUGE-L points, respectively. Compared with other strong baselines, our \textsc{Eit} can still show superiority on these datasets in terms of ROUGE-1 points, e.g., \textsc{Eit} surpasses \textsc{Surface}, \textsc{Dman}, \textsc{Bottom-up}, Talking-Head by 0.62, 0.64, 0.40 and 0.36 in terms of ROUGE-1 points, respectively. Notably, our \textsc{E-Eit} can achieve comparable performance with \textsc{Eit}.

\subsection{Language Modeling}
\label{sec:result_lm}

 Table \ref{tab:LM} presents the perplexity scores of various models on the WikiText-103 test set. Our \textsc{Eit} and \textsc{E-Eit} models outperform the baseline with PPL scores of 1.11 and 0.92, respectively. These results highlight the high expressiveness of our methods, as the improvements are achieved with only a negligible increase in parameters. Also, these results demonstrate the universality of our approach as applying our approach to the decoder side can also achieve improvement.

 \begin{table}[t!]
\renewcommand{\arraystretch}{1.25}
\centering
\small
\setlength{\tabcolsep}{5pt}
\resizebox{\linewidth}{!}{
\begin{tabular}{lccc}
\toprule
\textbf{Model}  & \bf Depth & $\bm{\theta}$~(M) & \bf Test PPL    \\
\midrule
Adaptive Transformer & 8 & 146.49 & 21.11 \\
\textsc{Eit} & 8 & 146.50 & \bf 20.00 \\
\textsc{E-Eit} & 8 & 146.49 &20.19 \\
\bottomrule
\end{tabular}
}
\caption{\label{tab:LM}
Results on the WikiText-103 dataset.
}
\end{table}
\begin{table}[t!]
\renewcommand{\arraystretch}{1.25}
\small
\centering
\setlength{\tabcolsep}{5pt}
\resizebox{\linewidth}{!}{
\begin{tabular}{llrrrr}
\toprule
\#&\textbf{Model} &
\textbf{En-De} & \textbf{En-Ro}
  \\

\midrule
  1& Transformer  &  27.13  &  34.23\\
   \midrule
  2& \textsc{Eit} & \bf 28.00 & \bf 35.10 \\
 3 & \quad -  Many-to-Many & 27.39& 34.71\\
 4&   \quad - Inner-Subspace Interaction & 25.79& 32.50\\
 5&   \quad - Cross-Subspace Interaction & 27.70& 34.53\\
\bottomrule
\end{tabular}
}
\caption{Ablation study on two tasks. }
\label{tab:ablation}
\end{table}

\section{Ablation Studies}
\label{sec:ablation}
\paragraph{Settings.} We gave detailed description about the settings of ablation studies. 
\begin{itemize}
    \item \textsc{Eit} - M2M: We directly applied ISI and CSI modules to the attention maps generated by Eq. (1). Notably, in both ISI and CSI, we maintain a consistent ratio between hidden size and input size, e.g., 2 and 8 for ISI and CSI, respectively, mirroring the EIT settings. 
    \item \textsc{Eit} - ISI: we directly applied CSI module to the $M^2$ attention maps generated by M2M. 
    \item \textsc{Eit} - CSI: we only applied ISI module to the $M^2$ attention maps generated by M2M. 
\end{itemize}

\paragraph{Results.}
Table \ref{tab:ablation} summarizes the impacts of removing each module on En-De and En-Ro tasks, respectively. First, we found removing any module~(or sub-module) results in obvious performance degradation (\#3,4,5 vs. \#2). These evidences indicate the indispensability of these modules.

Notably, when removing the M2M module (\#2 vs. \#3), we observe an obvious decline in performance on two translation tasks, indicating the importance of M2M module. Within our \textsc{Eit} framework, the M2M module, motivated by the \textit{complementary principle}, serves the critical purpose of supplying necessary information for subsequent interactions. Therefore, its absence impedes the effectiveness of our two interaction models. 

Furthermore, the omission of the ISI sub-module (\#2 vs. \#4) results in a significant and noticeable decrease in BLEU scores.  
One possible explanation is that while increasing the number of heads enhances the information capacity, it also introduces a certain degree of irrelevant information (noise) into the attention maps.
Consequently, a direct fusion of these heads fails to yield satisfactory outcomes. However, our \textsc{Eit} framework overcomes this challenge by incorporating the ISI sub-module, which provides an effective mechanism for discarding irrelevant information while retaining the benefits of the previous heads. This unique and innovative design sets our approach apart from the \textit{attention expansion} technique~\cite{Zhou-etal-2021-refiner}.

\input{Figure/Analysis/final_attention_maps_similarity}
\input{Figure/Analysis/representation_quality}

\section{Analysis on Attention Heads Behavior}
\subsection{\textsc{Eit} owns Higher Consensus}

As depicted in Figure \ref{fig:vis_final_map}, it is evident that \textsc{Eit} exhibits the highest average similarity among attention maps from various heads, surpassing all other models. This finding suggests that \textsc{Eit} demonstrates a greater consensus among attention heads. We attribute this achievement to the significant role played by M2M and dual-enhanced interaction. M2M facilitates the generation of rich information, while dual enhanced interaction efficiently leverages and refines the available information from different attention heads.

\paragraph{Discussions.} This phenomenon is contradictory to the findings of previous studies about head interaction~\cite{wang2022improved}. We speculate that this is because our interactions are more efficient, not only relying on an adequate number of attention heads but also operating in a hierarchical manner. These characteristics result in a consensus among the attention maps.

\begin{table}[t!]
\renewcommand{\arraystretch}{0.7}
\centering
\small
\setlength{\tabcolsep}{6pt}
\resizebox{0.9\linewidth}{!}{
\begin{tabular}{lrrrrrr}
\toprule
\multirow{2}{*}{\textbf{Model}} & \multicolumn{3}{c}{\bf Pruning Ratio}  \\
\cmidrule(r){2-4} 
& 0.0\%& 50.0\%& 87.5\% \\
\midrule
Transformer-48L &  29.60  & 27.64  & 1.86 \\
\textsc{Eit}-48L  & \bf 30.25 & \bf 29.09 & \bf 21.12\\
\bottomrule
\end{tabular}
}
\caption{BLEU scores of models with head pruning on the En-De task.}
\label{tab:bleu_with_head_prune}
\end{table}

\subsection{Benefits of High Consensus}
\paragraph{\textsc{Eit} Learns High-quality Representations}

We further investigate how consensus affects the layer representations. Following \citep{Gong2021VisionTW,pmlr-v139-dong21a,Shi-etal-2022-revising-over-smoothing-bert,WangZCW22-etal-antioversmooothing}, we adopt the 
token correlation $\mathcal{TC}$ to measure the quality of features~(the lower, the better). The token correlation is computed by the Pearson correlation coefficient~\citep{Benesty2009pcc}. 

Figure \ref{fig:token_correlation_measurement} exhibits the results on the test set of the En-De task. Notably, the features learned by \textsc{Eit} exhibit lower token correlation compared to the vanilla Transformer across all configurations. This indicates that \textsc{Eit} effectively learns improved layer representations.

Furthermore, we observe that the vanilla Transformer consistently maintains a relatively high token correlation from the first layer. This observation aligns with prior study~\cite{shleifer2022normformer}, suggesting that lower layers struggle to optimize effectively in pre-normalization Transformers. However, our \textsc{Eit} approach alleviates this issue.

\paragraph{\textsc{Eit} Makes Head Pruning Easier  }

To further explore the possibility of pruning the consensus attention maps, we introduce a simple head mask mechanism for head pruning during the inference phase as follows::
$
 \mathbf{O} = \sum_{i=1}^{M}\mathbf{\eta}_{i}\mathbf{A}^{i}\mathbf{X}\mathbf{W}_{V}^{i}\mathbf{W}_{O}^{i}
$, where $\mathbf{\eta}_{i} \in \{0,1\}$. Table \ref{tab:bleu_with_head_prune} exhibits the results on En-De tasks. Note that the head selection process is done in a straightforward manner, such as selecting heads by index, without considering their relative importance as highlighted in previous studies \citep{MichelLN19-etal-16-better-than-8}. Additionally, the head pruning operations are exclusively applied to the encoder side.
It is evident that \textsc{Eit} exhibits a high tolerance for head pruning without experiencing significant deterioration in performance. Such phenomenon sheds light on the research of head pruning and inference speeding.

\section{Analysis of Computational Efficiency}
\paragraph{MACs Comparison.}
Table \ref{tab:efficiency_analysis} displayed MACs comparison between \textsc{Eit} variants and transformer baselines. We can see that \textsc{Eit} can achieve an improvement of 0.87 BLEU points with an increase of just 0.1B MACs and 0.07M parameters. This indicated the efficiency of our \textsc{Eit} architecture. Besides, the efficient version of \textsc{Eit}. the \textsc{E-Eit} can achieve similar improvements with even fewer extra resource consumption.

\paragraph{Resource Comparison.} In addition to theoretically exploring the efficiency of \textsc{Eit} variants, we also measured the practical computational consumption during the training process. Without losing generality, we focused on the model for the \textit{base} configuration. We can see that \textsc{Eit} consumes 8\% more memory consumption and 45\% more
training costs than the baseline with a depth of 6. To mitigate this, we have proposed the \textsc{E-Eit} which only costs 5\%
more memory consumption and 10\% more training costs than the baseline but delivered similar performance compared to \textsc{Eit}. Notably, as shown in Table \ref{tab:en-de-en-ro}, the performance gap between \textsc{Eit} and \textsc{E-Eit} decreases as the model capacity increases.

\begin{table}[t!]
\renewcommand{\arraystretch}{1}
\small
\centering
\setlength{\tabcolsep}{4pt}
\resizebox{\linewidth}{!}{
\begin{tabular}{lrrrrr}
\toprule
\textbf{Model}  & \bf $\mathbf{\theta}$~(M)&
\bf MACs & \bf Time & \bf Memory &\bf BLEU
  \\

\midrule
   Transformer  & 61.56 & 10.0B & - & - &  27.13\\
   \textsc{Eit} & 61.63 &10.1B & 1.45$\times$ & 1.08$\times$ &\bf 28.00 \\
   \textsc{E-Eit} & 61.57 & 10.0B & 1.10$\times$ & 1.05$\times$& 27.72 \\
\bottomrule
\end{tabular}
}
\caption{\textbf{\textsc{Eit} variants are efficient as compared to transformers.} BLEU score is reported on the WMT’14 En-De dataset. We used 20 source and target tokens
for computing multiplication-addition operations
(MACs). }
\label{tab:efficiency_analysis}
\end{table}

\section{Related Work}
\paragraph{Low Bottleneck in Multi-Head Attention}
The ``Low Bottleneck'' issue in Multi-head Self-Attention (MHSA) occurs when adding more heads to Transformers, which does not correspondingly improve performance. \citet{pmlr-v119-bhojanapalli20a} first identified this issue, attributing it to the diminishing head dimension as the number of heads increases, which limits the creation of precise attention maps. In response, \citet{Shazeer-etal-2020} introduced ``talking-head attention,'' using two linear transformations around the SoftMax function to address this bottleneck. Later, \citet{Zhou-etal-2021-refiner} proposed a framework involving ``ghost heads'' to enrich attention patterns, differentiating from talking-head attention in the positioning of linear transformations and the number of ghost heads. Our approach introduces a many-to-many mapping in MHSA, using existing queries and keys for more attention maps through direct query-key multiplication.

\paragraph{Improved Multi-Head Mechanism}
Previous work has shown that multi-head attention can be further enhanced by encouraging individual attention heads to extract distinct information \citep{li-etal-2018-multi-head,cui-etal-2019-mixed,sukhbaatar-etal-2019-adaptive,Guo-etal-20,hao-etal-2019-multi}.
Another branch of research is designing more complex interactive modeling to make better use of the multiple subspace information~\citep{Shazeer-etal-2020,  Wang-etal-2020-Collaboration, li-etal-2019-information}. Besides, \citet{voita-etal-2019-analyzing} empirically demonstrates that some heads in attention are useless and can be pruned without performance degradation. Along this line, researchers investigate how to efficiently cut off redundant heads \citep{MichelLN19-etal-16-better-than-8,behnke-heafield-2020-losing}. Different from theirs, our study utilized the benefits of both diversity and consistency.

\section{Conclusions}
In this paper, we propose \textsc{Eit}, an alternative to the Transformer architecture. It further advances the multi-head schema by fully leveraging two principles in multi-view learning: the \textit{complementary principle} and the \textit{consensus principle}. In addition, \textsc{E-Eit} can serve as another choice considering the trade-off between performance and computation efficiency. Experimental results on four widely-used tasks demonstrate the effectiveness of \textsc{Eit}-variants, which deliver consistent improvements to the standard Transformer.

\section*{Acknowledgments}
This work was supported in part by the National Science Foundation of China (No.62276056), the Natural Science Foundation of Liaoning Province of China (2022-KF-16-01), the Fundamental Research Funds for the Central Universities (Nos. N2216016 and N2316002), the Yunnan Fundamental Research Projects (No. 202401BC070021), and the Program of Introducing Talents of Discipline to Universities, Plan 111 (No.B16009).

\section*{Limitations}
Besides the advantages endowed by \textsc{Eit}, there still exists a shortcoming that the computational efficiency of the group convolution cannot be satisfactory, although it is computationally efficient in theory. This is due to the lack of high-efficiency CUDA kernel support. We will release a more efficient optimization of group convolutions in the soon future.


\input{acl_latex.bbl}
\appendix
   
\newpage

\section{Detailed Setups of Experiments}
\label{Appendix:Experimental_Setups}

\subsection{Machine Translation Task}
\label{Appendix:Experimental_Setups_MT}

\paragraph{Dataset} We evaluated our approach on two widely used machine translation datasets: WMT'14 En-De and WMT'16 En-Ro. The En-De dataset contains approximately 4.5M tokenized training sentence pairs. We selected newstest2013 and newstest2014 as the validation and test data, respectively. As for the En-Ro dataset, it consists of 0.6M tokenized training sentence pairs. We performed shared BPE operations on both datasets to overcome the out-of-vocabulary (OOV) problem. Concretely, we set the size of BPE operations to 32K and 20K for En-De and En-Ro datasets, resulting in a shared vocabulary with sizes of 34040 and 19064, respectively. 

\paragraph{Model Configuration}  Our model architectures are based on Transformer~\citep{Vaswani-etal-2017}. We provided three basic configurations, namely \textit{base}, \textit{deep}, and \textit{big} which follow the configurations in \citet{Vaswani-etal-2017}. We adopted a pre-normalization strategy~\citep{wang-etal-2019-learning-deep} considering training stability under different configurations. The detailed settings of hyper-parameters are given in Table \ref{tab:config}.

\paragraph{Training \& Evaluation} Our implementations are based on Fairseq~\citep{ott-etal-2019-fairseq}.
Our experiments are performed on the GEFORCE RTX 3090 cards. We use 8 GEFORCE RTX 3090 cards to train models for the WMT'14 En-De task. As for the models on the WMT'16 En-Ro task, we train them on 4 GEFORCE RTX 3090 cards. The batch sizes for En-De and En-Ro tasks are 65536 and 16384, respectively. The total updates are 50K, 50K and 100K for \textit{base}, \textit{deep} and \textit{big} in En-De task, respectively.  We adopt Adam~\citep{Kingma-etal-2015} as an optimizer with an $\mathrm{adam}_{\beta}$ of (0.9, 0.997). The learning rate scheduler is \textit{invert sqrt} with a learning rate of 0.002 and warmup updates of 16000. We also adopt label smoothing with a ratio of 0.1 in all the experiments. More details are exhibited in Table \ref{tab:training}. During the evaluation process, we set the beam number to 4 and the length penalty to 0.6 for the En-De task. As for the En-Ro task, the number of beams is 5 and the length penalty is 1.3.

\begin{table*}[ht!]
\centering
\setlength{\tabcolsep}{4.5pt}
\begin{tabular}{lrrrcc}
\toprule
\multirow{2}{*}{\textbf{Dataset}}& \multicolumn{3}{c}{\textbf{Sentence}}& \multirow{2}{*}{\textbf{BPE}} & \multirow{2}{*}{\textbf{Vocab}}  \\
\cmidrule{2-4}
& \textbf{Train} & \textbf{Dev} & \textbf{Test} &  &  \\
\midrule
WMT'14 En-De & 4.5M&3.0K&3.0K & 32K & 34040\\
WMT'16 En-Ro  & 0.6M&2.0K& 2.0K &  20K&  19064\\
CNN/DailyMail &287K & 13.0K& 11.0K &  30K&  32584\\
CONLL &827K & 5.4K& 1.3K &  30K&  33136\\
WikiText-103 &103M & 218K& 246K &  -&  267740\\
\bottomrule
\end{tabular}
\caption{The details of datasets of language tasks.}
\label{tab:dataset_deatails}
\end{table*}

\begin{table*}[ht!]
\centering
\setlength{\tabcolsep}{2pt}
\begin{tabular}{llcrrrrrrrrr}
\toprule
\bf Task &{\bf Model} & {\bf Configuration}  & {$\mathbf{M}$} &{$\mathbf{M^H}$} &{$\mathbf{M^{H_{isi}}}$} &{$\mathbf{M^{H_{csi}}}$} &{$\mathbf{r}$} & {$\mathbf{K}_h^{isi}$} & {$\mathbf{K}_w^{isi}$} & {$\mathbf{K}_h^{csi}$} & {$\mathbf{K}_w^{csi}$} \\
\midrule
\multirow{6}{*}{MT}&\multirow{3}{*}{\textsc{Eit}} & \textit{base}  & 8 & -& 128 & 64 & 8 & 1 & 7 & 1 & 3\\
&&\textit{deep}  & 8 & -& 128 & 64 & 8 & 1 & 7 & 1 & 3\\
&&\textit{big}  & 16 & -& 256 & 256 & 16 & 1 & 7 & 1& 3\\
\cmidrule(r){2-12}
&\multirow{3}{*}{\textsc{E-Eit}}&\textit{base}  & 8 & 32& - & - & 8 & 1 & 7 & 1 & 7\\
&&\textit{deep}  & 8 & 32& - & - & 8 & 1 & 7 & 1 & 7\\
&&\textit{big} & 16 & 64& - & - & 16 & 1 & 7 & 1 & 7\\
\midrule
\multirow{2}{*}{AS}&\textsc{Eit} & \textit{base}  & 8 & -& 8 & 64 & 8 & 1 & 1 & 1 & 1\\
\cmidrule(r){2-12}
&\textsc{E-Eit}&\textit{base}  & 8 & 16& - & - & 8 & 1 & 1 & 1 & 1\\
\midrule
\multirow{2}{*}{GEC}&\textsc{Eit} & \textit{base}  & 8 & -& 128 & 128 & 8 & 1 & 7 & 1 & 3\\
\cmidrule(r){2-12}
&\textsc{E-Eit}&\textit{base}  & 8 & 64& - & - & 8 & 1 & 7 & 1 & 7\\
\midrule
\multirow{2}{*}{LM}&\textsc{Eit} & \textit{big}  & 8 & - & 64 & 32 & 8 & 1 & 1 & 1 & 1\\
\cmidrule(r){2-12}
&\textsc{E-Eit}&\textit{big}  & 8 & 8& - & - & 8 & 1 & 1 & 1 & 1\\
\toprule
\end{tabular}
\caption{The configurations of models on three sequence generation tasks. MT, AS, GEC and LM denote machine translation, abstractive summarization, grammar error correction and language modelling, respectively.}
\label{tab:config}
\end{table*}

\begin{table*}[ht!]
\centering
\small
\setlength{\tabcolsep}{2pt}
\begin{tabular}{lccccc}
\toprule
\bf Hyper-parameter& \bf WMT'14 En-De & \bf WMT'16 En-Ro & \bf CNN/DailyMail & \bf CONLL & \bf WikiText-103  \\
\midrule
GPUs & 8 & 4 & 8 & 8 & 8 \\
Batch & 4096 &4096 & 4096 & 4096 & 1024\\
UF & 2 & 1 & 4 & 2 & 8\\
Optimer & Adam  & Adam  & Adam  & Adam & Nag\\
{$\mathrm{{Adam}}_{\beta}$}  & (0.9, 0.997)  & (0.9, 0.997)  & (0.9, 0.997)  & (0.9, 0.980) & - \\
LR & 0.0020 &0.0020 & 0.0020 & 0.0015 & 0.0001\\
LR scheduler  & inverse sqrt &inverse sqrt & inverse sqrt & inverse sqrt & Cosine(t-mult=2)\\
Initial LR & 1$e^{-7}$ &1$e^{-7}$ & 1$e^{-7}$ & 1$e^{-7}$ & 1$e^{-7}$ \\
Total updates & 50K (100K) & 25K& 30K& 14K & 286K \\
Warmup updates & 16000 &8000 & 8000 & 4000 & 16000 \\
Weight decay & 0.0000 &0.0000 &  0.0001& 0.0001 & 0.0000\\
Label smoothing & 0.1 & 0.1 & 0.1 & 0.1 & 0.0\\
Dropout & 0.1 (0.3) & 0.1 (0.3) & 0.1 & 0.2 & 0.3 \\
Attention dropout & 0.1 & 0.1 & 0.1 & 0.1 & 0.1 \\
ReLU dropout & 0.1 & 0.1 & 0.1 & 0.1 & 0.1 \\
Word dropout & 0.0 &0.0 & 0.0 & 0.2 & 0.1\\
\toprule
\end{tabular}
\caption{The training setups of different tasks. UF denotes the update frequency of the gradient. (.) lists the values of hyper-parameters under the \textit{big} configuration, which vary from the values under the \textit{base} configuration. }
\label{tab:training}
\end{table*}

\subsection{Abstractive Summarization Task}
\paragraph{Dataset}
For abstractive summarization, we conduct experiments on a widely used corpus, e.g., CNN/DailyMail dataset. It consists of 287K training documents. Shared BPE operations with a size of 30K are performed on all the training data, resulting in a vocabulary of 32584.

\paragraph{Model Configuration}
We only provide the \textit{base} configuration of our \textsc{Eit} and \textsc{E-Eit} for abstractive summarization. The details are presented in Table \ref{tab:config}.

\paragraph{Training \& Evaluation}
We train models for an abstractive summarization task on 8 GEFORCE RTX 3090 cards with a batch size of 131072 and total updates of 30K. We adopt a weight decay strategy with a ratio of 0.0001. Other hyper-parameters are the same as that in machine translation tasks. You can find their settings in Table \ref{tab:training}. During testing, the number of beams is set to 4 and the length penalty is set to 2.0. Besides, we set the minimal length and maximum length to 55 and 140, respectively.  

\subsection{Grammar Error Correction Task}
\paragraph{Dataset}
For the grammar error correction task, we select the CONLL dataset to evaluate our approach. The CONLL dataset consists of 827K training sentences. We replicate the setup in \citet{Chollampatt-etal-2018-aaai} and adopt the word-level dropout technique~\citep{sennrich-etal-2016-edinburgh} to alleviate the overfitting problem. More details are listed in Table \ref{tab:dataset_deatails}.

\paragraph{Model Configuration}
For the grammar error correction task, we only provide the \textit{base} configuration of our \textsc{Eit} and \textsc{E-Eit}. The details are presented in Table \ref{tab:config}. Notice that the models on this task adopt a post-normalization strategy.

\paragraph{Training \& Evaluation}
We train models for the grammar error correction task on 8 GEFORCE RTX 3090 cards. The batch size is 65536 and the total updates are 14K. More training details are shown in Table \ref{tab:training}. During testing, the beams and length penalty are set to 6 and 0.6, respectively.

\subsection{Automatic Disease Diagnosis Task}
\paragraph{Dataset}
For the automatic disease diagnosis task, we select the ABIDE dataset to evaluate our approach. The ABIDE dataset consists of 1009 brain networks from 1009 real samples of 17 international sits. Due to the heterogeneity of this data, we adopt the shared data with re-standardized data splitting in \citet{kan2022brain}. Specifically, 70\%, 10\% and 20\% samples are served as the training, validation and test sets, respectively. 

\paragraph{Model Configuration}
For ABIDE task, we still follow the model configuration in \citet{kan2022brain}. Specifically, we build our BrainNet\textsc{Eit}F with a two-layer encoder. The number of heads $M$ is set to 4 for each layer. 

\paragraph{Training \& Evaluation}
We train all models including the BrainNetTF and BrainNet\textsc{Eit}F for 200 epochs on a single GEFORCE RTX 3090 card. Each model is trained by 5 times. We adopt Adam~\citep{Kingma-etal-2015} as an optimizer with an initial learning rate of $10^{-4}$ and a weight decay
of $10^{-4}$. The batch size is set to 64. We adopt the checkpoint of the final epoch for evaluating the test set.

\subsection{Language Modeling Task}
\paragraph{Dataset}
For the language modeling task, we select the WikiText-103 dataset to evaluate our approach. The training set consists of 103M words from 28K articles. While for the validation and test sets, they are made up of 218K and 246K words, respectively. In detail, we follow the instructions in Fairseq~\citep{ott-etal-2019-fairseq} to obtain and preprocess the data. The details are listed in Table \ref{tab:dataset_deatails}.

\paragraph{Model Configuration}
For WikiText-103 task, Both the baseline and our model are all 8-layer big models with 8 heads. Note that the baseline we adopted is adaptive input transformer~\citep{baevski2018adaptive}. In this task, the kernel sizes in DEI are all set to 1.

\paragraph{Training \& Evaluation}
The training and evaluation settings all follow the standard instructions for language modeling in PyTorch~\citep{ott-etal-2019-fairseq}. We train both baseline and \textsc{Eit} with 286000 updates. The details are given in Table \ref{tab:training}. As for the evaluation process, we adopt the checkpoint performing best on the validation set. We set the max-tokens, max-sentences, context-window to 3072, 1 and 2560, respectively.

\section{Details of Metrics}
\label{appendix:detail_metrics}
\subsection{Calculation of Head Distance}
Inspired by the attention metrics in \citet{Zhou-etal-2021-deepvit} and \citet{ WangZCW22-etal-antioversmooothing}, we measure the distance between different heads by calculating cosine similarity among attention maps. Notice that our metric focuses on the diversity of attention maps, which is quite different from them. Denote the dataset as $\mathcal{D}$, and the attention map of $h$-th head of $l$-th layer of $i$-th sample denotes as $\mathbf{A}^{(h,l,i)}$, the head similarity in $l$-th layer is computed by averaging the cosine similarity of every two heads in $i$-th layer across all samples as:

\begin{equation}
\begin{aligned}
\mathcal{HD}^{(l)} &= \frac{1}{|\mathcal{D}|} \frac{1}{M(M-1)}\frac{1}{T} \times \sum_{i=1}^{|\mathcal{D}|} \\
& \quad\left( \sum_{j=1}^{M}\sum_{k=1}^{M} \sum_{t=1}^{T} \mathrm{C}(\mathbf{A}^{(j,l,i)}_{t,:}, \mathbf{A}^{(k,l,i)}_{t,:}) - M \right)
\end{aligned}
\end{equation}

where $|\mathcal{D}|$ denotes the size of dataset, $M$ is the number of partition of features in attention, $T$ is the sequence length and C($\cdot$) denotes the cosine similarity function. We set $\mathcal{D}$ to the test set of the corresponding task. The obtained head similarity ranges from [0, 1]. The larger the head similarity, the lower the distances between different heads are.

\subsection{Calculation of Token Correlation}
We define a metric $\mathcal{TC}$, which measures the correlation among the representations of different tokens. Denote the dataset as $\mathcal{D}$, and the sequence representation of $i$-th sample in $l$-th layer denotes as $\mathbf{X}^{(l,i)}$, the token correlation of in $l$-th layer is computed as:
    

\begin{equation}
\begin{aligned} 
\mathcal{TC}^{(l)} =& \frac{1}{|\mathcal{D}|}\frac{1}{T(T-1)}\sum_{i=1}^{|\mathcal{D}|}\\&(\sum_{j=1}^{T} \sum_{k=1}^{T} \rho(\mathbf{X}_{j}^{(l,i)}, \mathbf{X}_{k}^{(l,i)}) -T)
 \end{aligned}
 \end{equation}

where $\rho(\cdot)$ denotes the pearson correlation function. Intuitively, the larger the $\mathcal{TC}$ is, the higher the token correlation is, degrading the model's learning capacity~\citep{Gong2021VisionTW}.

\subsection{Efficiency Comparison}
\label{subsec_efficiency_analysis}
\input{Figure/Analysis/efficient_comparision}
Despite the performance evaluation, memory consumption and computational cost are also two major concerns in the literature. Figure \ref{fig:efficient_compariosion_en_de} also displays the memory consumption and computational cost of models on the En-De task. \textsc{Eit} only costs 8.5\% more memory consumption and 44.4\% more training costs than the baseline with a depth of 6. However, the extra consumption goes larger as the depth goes deeper.

Besides, as aforementioned, we elaborately design an efficient version E-Eit that only costs 9.4\% more memory consumption and 21.7\% more training costs than the baseline under all the configurations on average. In this work, the many-to-many mapping rule is only applied on the encoder side. This is because the proposed M2M module and the subsequent ISI and CSI sub-modules will significantly enlarge the inference cost due to the heavy use of product attention on the decoder side, although it can attain further benefits in terms of BLEU.

\section{Visualization of Training and Validation Perplexity}

\input{Figure/Analysis/visual_ppl}

We plot the training and validation perplexity of Transformer and our \textsc{Eit} on the WMT'14 task in Figure \ref{fig:visual_ppl}. We can see that our \textsc{Eit} owns lower training and validation perplexity than Transformer.

\section{Hyper-Parameters Analysis (Kernel Size and Hidden Size)} 
\input{Figure/Analysis/parameter_analysis}

Since there are several hyper-parameters in both ISI and CSI sub-modules, it is necessary to figure out how they affect performance. Figure \ref{fig:parameter_analysis} (a-d) plots the performance of \textsc{Eit} against the kernel size and the hidden size. We can see that \textsc{Eit} can outperform Transformer in all choice of kernel size and hidden size. This observation can further help us trade off efficiency and performance well. For example, we can set CSI kernel size to 1 or ISI kernel size to 3 or $M^{H_{isi}}$ to $M^2$ or $M^{H_{csi}}$ to 4$M$ to own a more efficient \textsc{Eit}.

\section{Local Analysis}
\input{Figure/Analysis/local_analysis}

Local modeling is one of the widely accepted ways to improve the expressiveness of Transformer~\citep{yang-etal-2019-convolutional,fan-etal-2021-mask,li-2022-umst}.
In dual enhanced interaction, we apply convolution operations to attention maps, which has the potential to introduce local biases. To figure it out, we measure the localness of attention maps since if there is a local bias, each token will distribute larger attention weights on their neighboring tokens. We adopt the localness metric of \citet{fan-etal-2021-mask}, denoted as $\mathcal{C}$ (higher is better). More details are presented in the Appendix.

We plot the $\mathcal{C}$ value within a local region $w=0.1*T+1$, of models in En-De task and CNN-DailyMail task in Figure \ref{fig:localness_measure}. The value is computed over the test set. Due to the long sequence length, we only use a subset of the test set consisting of 1000 samples for the CNN-DailyMail task.  The results (mean) show no significant local enhancement phenomena in both tasks. Note that the attention maps in the first layer of \textsc{Eit} on the abstractive summarization have a strong local pattern, but the kernel sizes are set to 1 on this task.  So we conclude that the improvements do not come from local enhancement.
\begin{table}[]
\renewcommand{\arraystretch}{1}
\centering
\small
\setlength{\tabcolsep}{1.5pt}
\resizebox{\linewidth}{!}{
\begin{tabular}{lcccc}
\toprule
\textbf{Model} & \bf AUROC  & \bf  ACC &  \bf  SEN & \bf SPE  \\
\midrule
MvS-GCN~\citep{wen2022mvs}  & 69.0 &  69.4& 69.3&64.5\\
BrainNetTF~\citep{kan2022brain}  & 80.9±2.6 & 71.8±3.0   & 71.1±4.1  & 72.5±1.9\\
BrainNet\textsc{Eit}F  &  81.3±2.7 &  73.8±3.2  & \bf 73.9±5.8 &  75.6±4.7\\
BrainNet\textsc{E-Eit}F  &  \bf 82.9±3.3 & \bf 74.6±3.2  & \bf 72.2±5.3  & \bf 76.8±3.0\\
\bottomrule
\end{tabular}
}
\caption{\label{tab:brain}
AUROC, ACC, SEN and SPE points on ABIDE task.
}
\end{table}
\section{Evaluation on Automatic Brain Disease Diagnosis Task}
\label{sec:result_abdd}
We further inspect the potential of \textsc{Eit} to be severed as a general method beyond language tasks. The automatic brain disease diagnosis, a disease classification task that highly relies on precisely learning relationships among different brain regions has recently been dominated by graph convolution~\citep{wen2022mvs} and Transformer~\citep{kan2022brain}. We select a widely used real-world fMRI dataset: Autism Brain Imaging Data Exchange (ABIDE), which consists of 1009 brain networks from 17 international sites, of which 516 samples are autism spectrum disorder patients. We follow the preprocessing setup in \citet{kan2022brain} and adopt the CC200~\citep{craddock2012whole} as the Regin-of-Interest (ROI) partition template. We select two latest methods, the Mvs-GCN~\citep{wen2022mvs} and BrainNetTF~\citep{kan2022brain}, as our comparison. The
experimental setups and configurations of our BrainNet\textsc{Eit}F and BrainNet\textsc{EEit}F are the same as in \citet{kan2022brain}.
Each experiment is conducted 5 times and we report the mean and standard deviation of the four metrics: Accuracy (ACC), AUROC, Sensitivity (SEN) and Specificity (SPE).

\paragraph{Results}
We exhibit the ACC, AUROC, SEN and SPE of different models in Table \ref{tab:brain}. We can see that both BrainNet\textsc{Eit}F and BrainNet\textsc{EEit}F can outperform all the baselines in terms of all metrics. Similarly, thanks to the little increased parameters of our models, we can conclude that they have stronger expressiveness and can be easily extended to other scenarios.
\begin{table}[t!]
\renewcommand{\arraystretch}{0.4}
\centering
\small
\setlength{\tabcolsep}{5pt}
\resizebox{0.8\linewidth}{!}{
\begin{tabular}{cccccccc}
\toprule
\bf 1 & \bf 2 & \bf 3 & \bf 4 & \bf 5 & \bf 6 & \bf Time & \bf BLEU \\
\midrule
\checkmark & & & & & & 1.07$\times$ &27.76\\
 &  \checkmark & & & & & - &27.46 \\
 & &  \checkmark & & & & -& 27.40\\
 & & &  \checkmark & & & -& 27.38\\
 && & &  \checkmark &   & -&27.30\\
 &&& & &  \checkmark &   -&27.48 \\
 \midrule
 \checkmark & \checkmark& & & & & 1.13$\times$& \bf 28.08\\
 \checkmark & \checkmark& \checkmark& & & & 1.20$\times$ & 28.02 \\
 \checkmark & \checkmark& \checkmark& \checkmark & & & 1.27$\times$ & 28.05\\
 \checkmark & \checkmark& \checkmark& \checkmark & \checkmark& & 1.36$\times$ & 27.82 \\
 \checkmark & \checkmark& \checkmark& \checkmark & \checkmark&\checkmark & 1.45$\times$ & 28.00\\
 \bottomrule
\end{tabular}
}
\caption{\label{tab:effect_layer}
Layer evaluation of encoder with EMHA implementation. ``1" indicates the bottom layer.
}
\end{table}
\section{Further Analyses}
\subsection{Effect of Number of EIT Layers}
 Recent research \cite{shi-etal-2016-string, peters-etal-2018-deep, hao-etal-2019-multi} has demonstrated that various layers in the encoder of a model have a tendency to capture distinct syntax and semantic features. Consequently, each layer may have different requirements for promoting agreement among the representations. In light of this, we examine the impact of consensus on different layers. The results of the En-De task are presented in Tables \ref{tab:effect_layer}. The lowest layer clearly benefits from a higher degree of consensus compared to other layers, consistent with prior research \cite{shleifer2022normformer} indicating the challenges of optimizing shallow layers within the pre-normalization paradigm. However, by employing the consensus strategy, we enhance the learning of representations in shallow layers, giving them a significant advantage. Additionally, it is observed that incorporating consensus into a small subset of all layers can also yield good results, e.g., 28.08. These findings suggest two insights: 1) Our \textsc{Emha} is so powerful that can work well even only being applied to a small subset of all layers; and 2) more efficient utilization of consensus may achieve better performance while working more efficiently. 

\input{Figure/Analysis/dynamics_similarity}
 \subsection{Dynamics of Attention Map Similarity
during Computation}

Figure \ref{fig:dynamics_similarity} exhibits the dynamics of attention map similarity for the \textsc{Eit} 48L model on the En-De test set. The similarity between attention maps initially decreases and then increases as the dual interactions progress. This pattern is attributed to the two stages of our approach. In the ISI phase, interactions are modeled within each group instead of the whole, generating representative attention maps. As these groups operate independently, the similarity among these representatives is lower. Subsequently, in the CSI phase, interactions occur among these representatives, resulting in the final attention maps. This CSI enhances similarity among the attention maps, achieving the consensus.

\subsection{Back Translation Experiments}
In this section, we further investigate the effect of back-translation on our \textsc{Eit}. We conducted experiments on WMT'17 De-En and En-De to avoid misleading evaluation. Concretely, we randomly selected 7M monolingual German monolingual data and filtered the sentences by length (ones longer than 10 and shorter than 200 would be reserved). Then we used the already trained the WMT'17 De-En \textsc{Eit} model to generate the translations via top-k sampling (which is a more robust back-translation strategy than the original beam search. The beam is 4 and the length penalty is 1.5). We presented the results in Table \ref{tab:bt}.

When the back translation was applied, the vanilla Transformer and our \textsc{Eit} achieved BLEU scores of 29.53 and 30.39, respectively. This indicates that even under the back translation setting, our \textsc{Eit} can still achieve consistent performance improvements.

\begin{table}[t!]
\renewcommand{\arraystretch}{1}
\centering
\small
\setlength{\tabcolsep}{5pt}
\resizebox{0.8\linewidth}{!}{
\begin{tabular}{lc}
\toprule
\bf Model & \bf BLEU \\
\midrule
Transformer  & 28.59 \\
Transformer + BT & 29.53 (+ 0.94) \\
\midrule
\textsc{Eit} &  29.58 \\
\textsc{Eit} + BT &	30.39 (+ 0.81) \\
\bottomrule
\end{tabular}
}
\caption{\label{tab:bt}
Results of back translation.
}
\end{table}

\end{document}

%% file: Figure/many-to-many.tex
\definecolor{problue}{RGB}{93, 133, 159}
\definecolor{prodblue}{RGB}{19, 73, 108}
\definecolor{prored}{RGB}{216, 84, 62}
\definecolor{prolblue}{RGB}{130, 210, 198}
\definecolor{prored}{RGB}{ 159, 87, 87}
\definecolor{progrey}{RGB}{153,97,59}
\definecolor{proyellow}{RGB}{191, 153, 36}
\definecolor{prodgreen}{RGB}{4, 77, 21
}
\definecolor{prodarkgreen}{RGB}{10, 105, 121}
\definecolor{prodpurple}{RGB}{54, 23, 94}
\definecolor{prodred}{RGB}{96, 0, 0}

\definecolor{tiffanyblue}{RGB}{129,216,208}
\definecolor{Magenta2}{RGB}{238, 0, 238}
\definecolor{ugreen}{RGB}{118,218,145}
\definecolor{darkgreen}{RGB}{179, 214,110}
\definecolor{uyellow}{RGB}{239,202,102}
\definecolor{ublue}{RGB}{207,234,241}
\definecolor{upurple}{RGB}{196,165,222}
\definecolor{uorange}{RGB}{248,172,140}
\definecolor{upink}{RGB}{255,150,128}
\usetikzlibrary {patterns,patterns.meta}
\definecolor{palewhite}{RGB}{137, 157, 192}
\definecolor{bgreen}{RGB}{188,230,114}
\definecolor{jyellow}{RGB}{255,199,115}
\definecolor{lblue}{RGB}{135,206,250}
\definecolor{lpink}{RGB}{186,85,211}

\definecolor{dred}{RGB}{240,128,128}
\definecolor{dblue}{RGB}{65,105,225}
\definecolor{dgreen}{RGB}{60,179,113}
\definecolor{dpurple}{RGB}{153,50,204}

\begin{figure}[ht]
  \centering
    \tikzset{global scale/.style={
    scale=#1,
    every node/.append style={scale=#1}
  }
}
  
  \begin{tikzpicture}[global scale=0.75]

    \tikzstyle{circlenode}=[circle,minimum size=4pt,inner sep=0];

    \tikzstyle{Query}=[draw,rectangle,minimum height=1.5em,minimum width=2em,inner sep=0,thick, font=\footnotesize]
    
    \tikzstyle{ATTN}=[draw,rectangle,minimum height=1.4em,minimum width=1.4em,inner sep=0,thick, font=\footnotesize]
    
    \tikzstyle{ISI_boarder}=[rectangle,minimum height=2em,minimum width=10em,inner sep=0,thick, font=\footnotesize,rounded corners]

    
   \foreach \x/\j in {1/red,2/blue/,3/orange,4/tiffanyblue}
{
    \node[Query,fill=\j!30] (query\x) at (0 + 3em*\x ,0em) {$\bf{Q^{\x}}$};

    \node[Query,fill=\j!30] (key\x) at (0 + 3em*\x ,-3.5em) {$\bf{K^{\x}}$};
    
}

\path[fill=red, fill opacity=0.2] ([yshift=-0.8em]query1.center) -- ([xshift=-1.15em,yshift=0.9em]key1.center) -- ([xshift=-1.15em,yshift=-0.9em]key1.center) -- ([xshift=1.15em,yshift=-0.9em]key4.center) -- ([xshift=1.15em,yshift=0.9em]key4.center) -- ([yshift=-0.8em]query1.center);

\path[fill=blue, fill opacity=0.2] ([yshift=-0.8em]query2.center) -- ([xshift=-1.15em,yshift=0.9em]key1.center) -- ([xshift=-1.15em,yshift=-0.9em]key1.center) -- ([xshift=1.15em,yshift=-0.9em]key4.center) -- ([xshift=1.15em,yshift=0.9em]key4.center) -- ([yshift=-0.8em]query2.center);

\path[fill=orange, fill opacity=0.2] ([yshift=-0.8em]query3.center) -- ([xshift=-1.15em,yshift=0.9em]key1.center) -- ([xshift=-1.15em,yshift=-0.9em]key1.center) -- ([xshift=1.15em,yshift=-0.9em]key4.center) -- ([xshift=1.15em,yshift=0.9em]key4.center) -- ([yshift=-0.8em]query3.center);

\path[fill=tiffanyblue, fill opacity=0.2] ([yshift=-0.8em]query4.center) -- ([xshift=-1.15em,yshift=0.9em]key1.center) -- ([xshift=-1.15em,yshift=-0.9em]key1.center) -- ([xshift=1.15em,yshift=-0.9em]key4.center) -- ([xshift=1.15em,yshift=0.9em]key4.center) -- ([yshift=-0.8em]query4.center);
    
      \foreach \x in {1,...,4}
{

    \node[Query] (key\x) at (0 + 3em*\x ,-3.5em) {$\bf{K^{\x}}$};
    
}

      \foreach \x in {1,...,4}
{

    \node[ATTN] (attn1\x) at ([xshift=-2em-6.25em+2.5em*\x,yshift=4em]query1.center) {};
    
}

      \foreach \x in {1,...,4}
{

    \node[ATTN] (attn4\x) at ([xshift=2em-6.25em+2.5em*\x,yshift=4em]query4.center) {};
    
}
    
    \node[ISI_boarder,fill=red!30]  (group1) at ([xshift=1.25em]attn12.center) {};
    
     \node[ISI_boarder,fill=tiffanyblue!30]  (group4) at ([xshift=1.25em]attn42.center) {};
    

      \foreach \x in {1,...,4}
{

    \node[ATTN] (attn1\x) at ([xshift=-2em-6.25em+2.5em*\x,yshift=4em]query1.center) {$\bf{S^{\x}}$};
    
}
\foreach \x/\y in {1/13,2/14,3/15,4/16}
{

    \node[ATTN] (attn4\x) at ([xshift=2em-6.25em+2.5em*\x,yshift=4em]query4.center) {$\bf{S^{\y}}$};
    
}

\node[] (quesheng) at ([yshift=4em,xshift=1.5em]query2.center) {$\dots$};

\draw[thick,->] (query1) -- ([yshift=1.5em]query1.center) -| (group1);

\draw[thick,->] (query4) -- ([yshift=1.5em]query4.center) -| (group4);

\draw[thick,->] (query2) -- ([yshift=1.5em]query2.center) -| (quesheng);

\draw[thick,->] (query3) -- ([yshift=1.5em]query3.center) -| (quesheng);

\node[] at ([yshift=1.5em]quesheng.center) {\textbf{Generated Attention Maps: $16$}};


\node[left] at ([xshift=-2em]query1.center) {\textbf{Queries}};

\node[left] at ([xshift=-2em]key1.center) {\textbf{Keys}};

\node[left] at ([xshift=-2em, yshift=1.75em]key1.center) {\textbf{Mappings}};

  \end{tikzpicture}
  \caption{The illustration of many-to-many mapping scheme ($M=4$). 
  }
  \label{fig:architecture}
\end{figure}
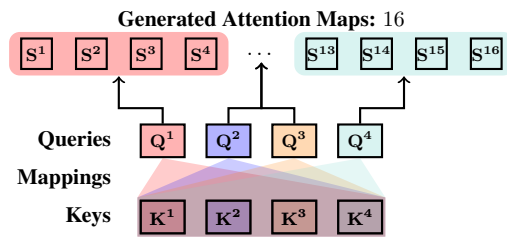

%% file: Figure/dual_enhanced_interaction.tex
\definecolor{problue}{RGB}{93, 133, 159}
\definecolor{prodblue}{RGB}{19, 73, 108}
\definecolor{prored}{RGB}{216, 84, 62}
\definecolor{prolblue}{RGB}{130, 210, 198}
\definecolor{prored}{RGB}{ 159, 87, 87}
\definecolor{progrey}{RGB}{153,97,59}
\definecolor{proyellow}{RGB}{191, 153, 36}
\definecolor{prodgreen}{RGB}{4, 77, 21
}
\definecolor{prodarkgreen}{RGB}{10, 105, 121}
\definecolor{prodpurple}{RGB}{54, 23, 94}
\definecolor{prodred}{RGB}{96, 0, 0}

\definecolor{tiffanyblue}{RGB}{129,216,208}
\definecolor{Magenta2}{RGB}{238, 0, 238}
\definecolor{ugreen}{RGB}{118,218,145}
\definecolor{darkgreen}{RGB}{179, 214,110}
\definecolor{uyellow}{RGB}{239,202,102}
\definecolor{ublue}{RGB}{207,234,241}
\definecolor{upurple}{RGB}{196,165,222}
\definecolor{uorange}{RGB}{248,172,140}
\definecolor{upink}{RGB}{255,150,128}
\usetikzlibrary {patterns,patterns.meta}
\definecolor{palewhite}{RGB}{137, 157, 192}
\definecolor{bgreen}{RGB}{188,230,114}
\definecolor{jyellow}{RGB}{255,199,115}
\definecolor{lblue}{RGB}{135,206,250}
\definecolor{lpink}{RGB}{186,85,211}

\definecolor{dred}{RGB}{240,128,128}
\definecolor{dblue}{RGB}{65,105,225}
\definecolor{dgreen}{RGB}{60,179,113}
\definecolor{dpurple}{RGB}{153,50,204}

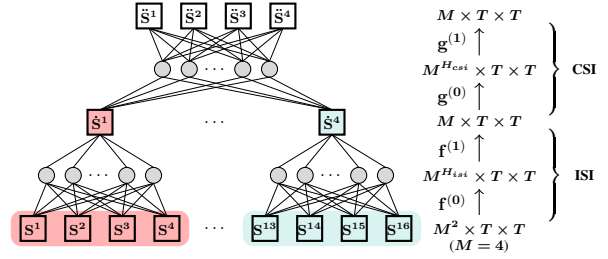
\begin{figure}[]
  \centering
    \tikzset{global scale/.style={
    scale=#1,
    every node/.append style={scale=#1}
  }
}
  
  \begin{tikzpicture}[global scale=0.61]

    \tikzstyle{circlenode}=[draw, circle,minimum size=0.9em,inner sep=0];

    \tikzstyle{Query}=[draw,rectangle,minimum height=1.5em,minimum width=2em,inner sep=0,thick, font=\footnotesize]
    
    \tikzstyle{ATTN}=[draw,rectangle,minimum height=1.4em,minimum width=1.4em,inner sep=0,thick, font=\footnotesize]
    
    \tikzstyle{ISI_boarder}=[rectangle,minimum height=2em,minimum width=10em,inner sep=0,thick, font=\footnotesize,rounded corners]

    
   \foreach \x/\j in {1/red,2/blue/,3/orange,4/tiffanyblue}
{
    \node[] (query\x) at (0 + 3em*\x ,0em) {};
    
}

      \foreach \x in {1,...,4}
{

    \node[ATTN] (attn1\x) at ([xshift=-2em-6.25em+2.5em*\x]query1.center) {};
    
}

      \foreach \x in {1,...,4}
{

    \node[ATTN] (attn4\x) at ([xshift=2em-6.25em+2.5em*\x]query4.center) {};
    
}
    
    \node[ISI_boarder,fill=red!30]  (group1) at ([xshift=1.25em]attn12.center) {};
    
     \node[ISI_boarder,fill=tiffanyblue!30]  (group4) at ([xshift=1.25em]attn42.center) {};
    

      \foreach \x in {1,...,4}
{

    \node[ATTN] (attn1\x) at ([xshift=-2em-6.25em+2.5em*\x]query1.center) {$\bf{S^{\x}}$};
    
}
\foreach \x/\y in {1/13,2/14,3/15,4/16}
{

    \node[ATTN] (attn4\x) at ([xshift=2em-6.25em+2.5em*\x]query4.center) {$\bf{S^{\y}}$};
    
}

\node[] (quesheng) at ([xshift=1.5em]query2.center) {$\dots$};

\node[circlenode,fill=gray!30] (hidden11) at ([yshift=3em,xshift=-3em]group1.center) {};

\node[circlenode,fill=gray!30] (hidden12) at ([yshift=3em,xshift=-1.5em]group1.center) {};

\node[circlenode,fill=gray!30] (hidden13) at ([yshift=3em,xshift=1.5em]group1.center) {};

\node[circlenode,fill=gray!30] (hidden14) at ([yshift=3em,xshift=3em]group1.center) {};

\node[] (quesheng_hidden1) at ([yshift=3em]group1.center) {$\dots$};

\foreach \x in {1,2,3,4}
{
    \foreach \y in {1,2,3,4} {

    \draw[very thin] ([yshift=-0.45em]hidden1\x.center) -- ([yshift=0.7em]attn1\y.center);
    }
}

\node[ATTN,fill=red!30] (attn1) at ([yshift=3em]quesheng_hidden1.center) {$\bf{\dot{ S}^{1}}$};

\foreach \x in {1,2,3,4}
{
    \draw[very thin] ([yshift=0.45em]hidden1\x.center) -- ([yshift=-0.7em]attn1.center);

}

\foreach \x in {1,2,3,4}
{
    \node[circlenode,fill=gray!30]  (hidden4\x) at ([xshift=13em]hidden1\x.center) {};

}

\node[] (quesheng_hidden4) at ([yshift=3em]group4.center) {$\dots$};

\foreach \x in {1,2,3,4}
{
    \foreach \y in {1,2,3,4} {

    \draw[very thin] ([yshift=-0.45em]hidden4\x.center) -- ([yshift=0.7em]attn4\y.center);
    }
}
  
 \node[ATTN, fill=tiffanyblue!30] (attn4) at ([yshift=3em]quesheng_hidden4.center) {$\bf{\dot{ S}^{4}}$};

\foreach \x in {1,2,3,4}
{
    \draw[very thin] ([yshift=0.45em]hidden4\x.center) -- ([yshift=-0.7em]attn4.center);

}

\node[] (quesheng_isi_output) at ([yshift=6em]quesheng.center) {$\dots$};

\node[circlenode,fill=gray!30] (hidden1) at ([yshift=3em,xshift=-3em]quesheng_isi_output.center) {};

\node[circlenode,fill=gray!30] (hidden2) at ([yshift=3em,xshift=-1.5em]quesheng_isi_output.center) {};

\node[circlenode,fill=gray!30] (hidden3) at ([yshift=3em,xshift=1.5em]quesheng_isi_output.center) {};

\node[circlenode,fill=gray!30] (hidden4) at ([yshift=3em,xshift=3em]quesheng_isi_output.center) {};

\node[] (quesheng_csi_hidden) at ([yshift=3em]quesheng_isi_output.center) {$\dots$};

\foreach \x in {1,4}
{
    \foreach \y in {1,2,3,4}{

    \draw[very thin] ([yshift=0.7em]attn\x.center) -- ([yshift=-0.45em]hidden\y.center);
    }
}

\foreach \x/\y in {1/13,2/14,3/15,4/16}
{

    \node[ATTN] (attn_final\x) at ([xshift=-6.25em+2.5em*\x,yshift=3em]quesheng_csi_hidden.center) {$\bf{\ddot{S}^{\x}}$};
    
}

\foreach \x in {1,2,3,4}
{
    \foreach \y in {1,2,3,4}{

    \draw[very thin] ([yshift=0.45em]hidden\x.center) -- ([yshift=-0.7em]attn_final\y.center);
    }
}

    
        
         

    
    \node[font=\footnotesize] (shape1) at ([xshift=4.5em]attn44.center) {\bm{${M^2 \times T \times T}$}};
    
    \node[font=\footnotesize]  at ([yshift=-1em]shape1.center) {(\bm{$M=4$})};
    
    \node[] at ([xshift=-1.5em,yshift=1.5em]shape1.center) {$\bf f^{(0)}$};
    
    \node[font=\footnotesize] (shape2) at ([xshift=4.5em,yshift=3em]attn44.center) {\bm{${M^{H_{isi}} \times T \times T}$}};
    
    \node[] at ([xshift=-1.5em,yshift=1.5em]shape2.center) {$\bf f^{(1)}$};
    
    \node[font=\footnotesize] (shape3) at ([xshift=4.5em,yshift=6em]attn44.center) {\bm{${M \times T \times T}$}};
    
    \node[] at ([xshift=-1.5em,yshift=1.5em]shape3.center) {$\bf g^{(0)}$};
     \node[font=\footnotesize](shape4) at ([xshift=4.5em,yshift=9em]attn44.center) {\bm{${M^{H_{csi}} \times T \times T}$}};
     
     \node[] at ([xshift=-1.5em,yshift=1.5em]shape4.center) {$\bf g^{(1)}$};
     
     \node[font=\footnotesize] (shape5)at ([xshift=4.5em,yshift=12em]attn44.center) {\bm{${M \times T \times T}$}};
     \draw[->] (shape1) -- (shape2);
      \draw[->] (shape2) -- (shape3);
       \draw[->] (shape3) -- (shape4);
        \draw[->] (shape4) -- (shape5);

\node[rotate = 90] at ([xshift=4.2em]shape2.center) {$\underbrace{\hspace{2cm}}$};

\node[rotate = 90] at ([xshift=4.2em]shape4.center) {$\underbrace{\hspace{2cm}}$};

\node[align=center, font=\footnotesize] at ([xshift=5.9em]shape2.center) {\bf ISI};

\node[align=center, font=\footnotesize] at ([xshift=5.9em]shape4.center) {\bf CSI};

  \end{tikzpicture}
  \vskip -0.1in
  \caption{Illustration of dual enhanced interaction in \textsc{Eit} ($M=4$). We omit the ReLU for simplicity.
  }
  \label{fig:dei}
\end{figure}

%% file: Figure/eeit.tex
\definecolor{problue}{RGB}{93, 133, 159}
\definecolor{prodblue}{RGB}{19, 73, 108}
\definecolor{prored}{RGB}{216, 84, 62}
\definecolor{prolblue}{RGB}{130, 210, 198}
\definecolor{prored}{RGB}{ 159, 87, 87}
\definecolor{progrey}{RGB}{153,97,59}
\definecolor{proyellow}{RGB}{191, 153, 36}
\definecolor{prodgreen}{RGB}{4, 77, 21
}
\definecolor{prodarkgreen}{RGB}{10, 105, 121}
\definecolor{prodpurple}{RGB}{54, 23, 94}
\definecolor{prodred}{RGB}{96, 0, 0}

\definecolor{tiffanyblue}{RGB}{129,216,208}
\definecolor{Magenta2}{RGB}{238, 0, 238}
\definecolor{ugreen}{RGB}{118,218,145}
\definecolor{darkgreen}{RGB}{179, 214,110}
\definecolor{uyellow}{RGB}{239,202,102}
\definecolor{ublue}{RGB}{207,234,241}
\definecolor{upurple}{RGB}{196,165,222}
\definecolor{uorange}{RGB}{248,172,140}
\definecolor{upink}{RGB}{255,150,128}
\usetikzlibrary {patterns,patterns.meta}
\definecolor{palewhite}{RGB}{137, 157, 192}
\definecolor{bgreen}{RGB}{188,230,114}
\definecolor{jyellow}{RGB}{255,199,115}
\definecolor{lblue}{RGB}{135,206,250}
\definecolor{lpink}{RGB}{186,85,211}

\definecolor{dred}{RGB}{240,128,128}
\definecolor{dblue}{RGB}{65,105,225}
\definecolor{dgreen}{RGB}{60,179,113}
\definecolor{dpurple}{RGB}{153,50,204}

\begin{figure}[ht!]
  \centering
    \tikzset{global scale/.style={
    scale=#1,
    every node/.append style={scale=#1}
  }
}
  
  \begin{tikzpicture}[global scale=0.61]

    \tikzstyle{circlenode}=[draw, circle,minimum size=0.9em,inner sep=0];

    \tikzstyle{Query}=[draw,rectangle,minimum height=1.5em,minimum width=2em,inner sep=0,thick, font=\footnotesize]
    
    \tikzstyle{ATTN}=[draw,rectangle,minimum height=1.4em,minimum width=1.4em,inner sep=0,thick, font=\footnotesize]
    
    \tikzstyle{ISI_boarder}=[rectangle,minimum height=2em,minimum width=10em,inner sep=0,thick, font=\footnotesize,rounded corners]

    
   \foreach \x/\j in {1/red,2/blue/,3/orange,4/tiffanyblue}
{
    \node[] (query\x) at (0 + 3em*\x ,0em) {};
    
}

      \foreach \x in {1,...,4}
{

    \node[ATTN] (attn1\x) at ([xshift=-2em-6.25em+2.5em*\x]query1.center) {};
    
}

      \foreach \x in {1,...,4}
{

    \node[ATTN] (attn4\x) at ([xshift=2em-6.25em+2.5em*\x]query4.center) {};
    
}
    
    \node[ISI_boarder,fill=red!30]  (group1) at ([xshift=1.25em]attn12.center) {};
    
     \node[ISI_boarder,fill=tiffanyblue!30]  (group4) at ([xshift=1.25em]attn42.center) {};
    

      \foreach \x in {1,...,4}
{

    \node[ATTN] (attn1\x) at ([xshift=-2em-6.25em+2.5em*\x]query1.center) {$\bf{S^{\x}}$};
    
}
\foreach \x/\y in {1/13,2/14,3/15,4/16}
{

    \node[ATTN] (attn4\x) at ([xshift=2em-6.25em+2.5em*\x]query4.center) {$\bf{S^{\y}}$};
    
}

\node[] (quesheng) at ([xshift=1.5em]query2.center) {$\dots$};

\node[circlenode,fill=gray!30] (hidden11) at ([yshift=3em,xshift=-3em]group1.center) {};

\node[circlenode,fill=gray!30] (hidden12) at ([yshift=3em,xshift=-1.5em]group1.center) {};

\node[circlenode,fill=gray!30] (hidden13) at ([yshift=3em,xshift=1.5em]group1.center) {};

\node[circlenode,fill=gray!30] (hidden14) at ([yshift=3em,xshift=3em]group1.center) {};

\node[] (quesheng_hidden1) at ([yshift=3em]group1.center) {$\dots$};

\foreach \x in {1,2,3,4}
{
    \foreach \y in {1,2,3,4} {

    \draw[very thin] ([yshift=-0.45em]hidden1\x.center) -- ([yshift=0.7em]attn1\y.center);
    }
}

\node[] (attn1) at ([yshift=0em]quesheng_hidden1.center) {};

\foreach \x in {1,2,3,4}
{
    \node[circlenode,fill=gray!30]  (hidden4\x) at ([xshift=13em]hidden1\x.center) {};

}

\node[] (quesheng_hidden4) at ([yshift=3em]group4.center) {$\dots$};

\foreach \x in {1,2,3,4}
{
    \foreach \y in {1,2,3,4} {

    \draw[very thin] ([yshift=-0.45em]hidden4\x.center) -- ([yshift=0.7em]attn4\y.center);
    }
}
  
 \node[] (attn4) at ([yshift=0em]quesheng_hidden4.center) {};

\node[] (quesheng_isi_output) at ([yshift=6em]quesheng.center) {};

\node[] (quesheng_csi_hidden) at ([yshift=-3em]quesheng_isi_output.center) {$\dots$};

\foreach \x/\y in {1/13,2/14,3/15,4/16}
{

    \node[ATTN] (attn_final\x) at ([xshift=-6.25em+2.5em*\x,yshift=3em]quesheng_csi_hidden.center) {$\bf{\ddot{S}^{\x}}$};
    
}

    \node[font=\footnotesize] (shape1) at ([xshift=4.5em]attn44.center) {\bm{${M^2 \times T \times T}$}};
    
    \node[font=\footnotesize]  at ([yshift=-1em]shape1.center) {(\bm{$M=4$})};
    
    \node[] at ([xshift=-1.5em,yshift=1.5em]shape1.center) {$\bf f^{(0)}$};
    
    \node[font=\footnotesize] (shape2) at ([xshift=4.5em,yshift=3em]attn44.center) {\bm{${M^{H} \times T \times T}$}};
    
    \node[] at ([xshift=-1.5em,yshift=1.5em]shape2.center) {$\bf g^{(0)}$};
    
    \node[font=\footnotesize] (shape3) at ([xshift=4.5em,yshift=6em]attn44.center) {\bm{${M \times T \times T}$}};

     \draw[->] (shape1) -- (shape2);
      \draw[->] (shape2) -- (shape3);

\foreach \x in {1,2,3,4}
{
    \foreach \y in {1,2,3,4} {

    \draw[very thin] ([yshift=0.45em]hidden4\x.center) -- ([yshift=-0.7em]attn_final\y.center);
    }
}

\foreach \x in {1,2,3,4}
{
    \foreach \y in {1,2,3,4} {

    \draw[very thin] ([yshift=0.45em]hidden1\x.center) -- ([yshift=-0.7em]attn_final\y.center);
    }
}

\node[rotate = 90] at ([xshift=4.2em,yshift=-1.7em]shape2.center) {$\underbrace{\hspace{1.1cm}}$};

\node[align=center, font=\footnotesize] at ([xshift=5.9em,yshift=-1.7em]shape2.center) {\bf ISI };

\node[rotate = 90] at ([xshift=4.2em,yshift=-1.3em]shape3.center) {$\underbrace{\hspace{1.1cm}}$};

\node[align=center, font=\footnotesize] at ([xshift=5.9em,yshift=-1.3em]shape3.center) {\bf CSI };

  \end{tikzpicture}
  \vskip 0.1in
  \caption{Illustration of dual enhanced interaction in efficient \textsc{Eit} ($M=4$). We omit the ReLU for simplicity.
  }
  \label{fig:eeit}
\end{figure}
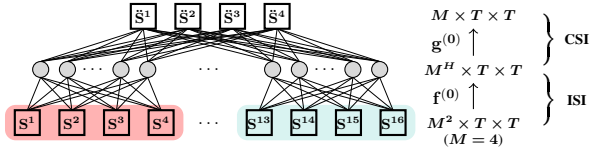

%% file: Figure/Analysis/final_attention_maps_similarity.tex
\definecolor{tiffanyblue}{RGB}{129,216,208}
\definecolor{bangdiblue}{RGB}{0,149,182}
\definecolor{kleinblue}{RGB}{0,47,167}
\definecolor{kabuliblue}{RGB}{26,85,153}
\definecolor{purple}{RGB}{138,43,226}
\usetikzlibrary{matrix}
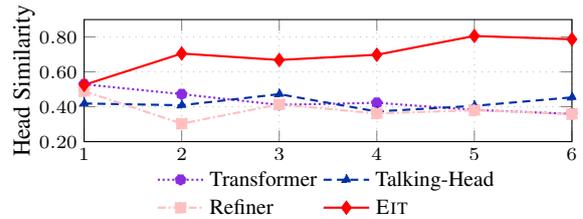
\begin{figure}[t!]
  \centering
   \tikzset{global scale/.style={
    scale=#1,
    every node/.append style={scale=#1}
  }
}
  \begin{tikzpicture}[global scale=1]

  \pgfplotsset{set layers}

     	\scriptsize{
      \begin{axis}[
	 at={(0,0)},
      ymajorgrids,
      xmajorgrids,
      grid style=dotted,
      width=0.5\textwidth,
      height=.2\textwidth,
      legend style={at={(0.5,-0.15)}, anchor=north, legend columns=2},
      xlabel={\small{Layer Index}},
      ylabel={\small{Head Similarity}},
      ylabel style={yshift=-1.8em},xlabel style={yshift=1.0em},
      yticklabel style={/pgf/number format/precision=2,/pgf/number format/fixed zerofill},
      ymin=0.2,ymax=0.9, ytick={0.2, 0.4, 0.6, 0.8},
      xmin=1,xmax=6,xtick={1,2,3,4,5,6},
      legend style={draw=none,yshift=0em,xshift=0em, font={\small},cells={anchor=west}},     
      ]
     
      \addplot[purple, mark=*,mark size=2pt,thick,densely dotted,mark options={fill=purple,draw=purple,line width=0.01pt}] coordinates { 
       (1, 0.5289) (2,0.4724) (3,0.4109) (4,0.4228) (5,0.3820) (6,0.3592) 
      };
      \addlegendentry{\scalebox{0.9}{Transformer}}
      
      \addplot[kleinblue, mark=triangle*,mark size=2pt,thick,densely dashed,mark options={fill=kleinblue,draw=kleinblue,line width=0.01pt}] coordinates { 
    (1, 0.4189) (2,0.4087) (3,0.4724) (4,0.3729) (5,0.4052) (6,0.4539) 
      };
      \addlegendentry{\scalebox{0.9}{Talking-Head}}

      \addplot[pink, dotted,mark=square*,mark size=2pt,thick,densely dashdotted,mark options={fill=pink,draw=pink,line width=0.01pt}] coordinates {
      (1, 0.4897) (2,0.3033) (3,0.4136) (4,0.3617) (5,0.3803) (6,0.3581) 
      };
      \addlegendentry{\scalebox{0.9}{Refiner}}
      
      \addplot[red, mark=diamond*,mark size=2.5pt,thick,solid,mark options={fill=red,draw=red,line width=0.01pt}] coordinates { 
       (1, 0.5253) (2,0.7052) (3,0.6682) (4,0.6982) (5,0.8059) (6,0.7871) 
      };
      \addlegendentry{\scalebox{0.9}{\textsc{Eit}}}

      \end{axis}
     }
  \end{tikzpicture}
  \vskip -0.1in
  \caption{Cosine similarity among attention maps of different models on En-De task.}\label{fig:vis_final_map}
\end{figure}

%% file: Figure/Analysis/representation_quality.tex
\definecolor{tiffanyblue}{RGB}{129,216,208}
\definecolor{bangdiblue}{RGB}{0,149,182}
\definecolor{kleinblue}{RGB}{0,47,167}
\definecolor{kabuliblue}{RGB}{26,85,153}
\definecolor{purple}{RGB}{138,43,226}

\usetikzlibrary{matrix}

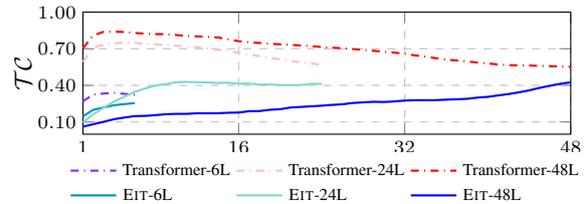
\begin{figure}[t!]
  \centering
  \tikzset{global scale/.style={
    scale=#1,
    every node/.append style={scale=#1}
  }}

  \begin{tikzpicture}[global scale=1]

    \pgfplotsset{set layers}

    \scriptsize{
      \begin{axis}[
        at={(0\textwidth,0)},
        ymajorgrids,
        xmajorgrids,
        grid style=dashed,
        width=0.5\textwidth,
        height=.2\textwidth,
        legend style={at={(0.5,-0.15)}, anchor=north, legend columns=3},
        xlabel={\small{Layer Index}},
        ylabel={\small{$\mathcal{TC}$}},
        ylabel style={yshift=-1.8em}, xlabel style={yshift=1.0em},
        yticklabel style={/pgf/number format/precision=2,/pgf/number format/fixed zerofill},
        ymin=0, ymax=1, ytick={0.1, 0.4, 0.7, 1.0},
        xmin=1, xmax=48, xtick={1, 16, 32, 48},
        legend style={draw=none, yshift=0em, xshift=0em, font={\small}, cells={anchor=west}},
      ]

      \addplot[dash dot, purple, mark=none, mark size=1pt, thick, mark options={solid}] coordinates {
        (1, 0.2699) (2, 0.3249) (3, 0.3332) (4, 0.3383) (5, 0.3303) (6, 0.3222)
      };
      \addlegendentry{\scalebox{.7}{Transformer-6L}}

    \addplot[dash dot, pink, mark=none, mark size=1pt, thick, mark options={solid}] coordinates {
        (1, 0.5931236147880554) (2, 0.7095480561256409) (3, 0.7335400581359863) (4, 0.74444580078125) (5, 0.7505544424057007) (6, 0.7492772936820984) (7, 0.739399254322052) (8, 0.7366998791694641) (9, 0.730485737323761) (10, 0.7251790761947632) (11, 0.7189871072769165) (12, 0.7127450108528137) (13, 0.7027862668037415) (14, 0.6921385526657104) (15, 0.6818290948867798) (16, 0.6662127375602722) (17, 0.6521402597427368) (18, 0.6395431756973267) (19, 0.6267321109771729) (20, 0.6111516356468201) (21, 0.599248468875885) (22, 0.5887652039527893) (23, 0.5800646543502808) (24, 0.5673767924308777)
      };
      \addlegendentry{\scalebox{.7}{Transformer-24L}}

         \addplot[dash dot, red, mark=none, mark size=1pt, thick, mark options={solid}] coordinates {
        (1, 0.7050339579582214) (2, 0.8091227412223816) (3, 0.8393278121948242) (4, 0.8399634957313538) (5, 0.836832582950592) (6, 0.8296874761581421) (7, 0.8229408860206604) (8, 0.81984543800354) (9, 0.8197107911109924) (10, 0.812329888343811) (11, 0.8046137690544128) (12, 0.801480233669281) (13, 0.7971142530441284) (14, 0.7923195362091064) (15, 0.7773875594139099) (16, 0.7617650628089905) (17, 0.7529897689819336) (18, 0.7474364042282104) (19, 0.7407044172286987) (20, 0.7365961670875549) (21, 0.7313146591186523) (22, 0.7270938754081726) (23, 0.7212213277816772) (24, 0.7141133546829224) (25, 0.7084481716156006) (26, 0.7040209174156189) (27, 0.6975299715995789) (28, 0.6904275417327881) (29, 0.6831967234611511) (30, 0.6744709014892578) (31, 0.6685528755187988) (32, 0.6589681506156921) (33, 0.645439863204956) (34, 0.6352242827415466) (35, 0.6225499510765076) (36, 0.610893189907074) (37, 0.6038575768470764) (38, 0.5962718725204468) (39, 0.5868430137634277) (40, 0.5796791911125183) (41, 0.5713546276092529) (42, 0.5661618113517761) (43, 0.562894344329834) (44, 0.5628546476364136) (45, 0.5597568154335022) (46, 0.5573935508728027) (47, 0.5550998449325562) (48, 0.5516771674156189)
      };
      \addlegendentry{\scalebox{.7}{Transformer-48L}}
    
      \addplot[thick, bangdiblue, mark=none, mark size=1pt, thick, mark options={solid}] coordinates {
        (1, 0.1466) (2, 0.2003) (3, 0.2190) (4, 0.2379) (5, 0.2462) (6, 0.2551)
      };
      \addlegendentry{\scalebox{.7}{\textsc{Eit}-6L}}

      \addplot[thick, tiffanyblue, mark=none, mark size=1pt, thick, mark options={solid}] coordinates {
        (1, 0.09632255882024765) (2, 0.16442978382110596) (3, 0.2170943170785904) (4, 0.2675982713699341) (5, 0.3075869381427765) (6, 0.3469473421573639) (7, 0.3747269809246063) (8, 0.39891132712364197) (9, 0.4106798768043518) (10, 0.4257860779762268) (11, 0.4285912811756134) (12, 0.42594748735427856) (13, 0.4239555895328522) (14, 0.41984039545059204) (15, 0.42181235551834106) (16, 0.41319116950035095) (17, 0.4153578579425812) (18, 0.40616974234580994) (19, 0.407535195350647) (20, 0.4022296369075775) (21, 0.40237393975257874) (22, 0.40233615040779114) (23, 0.4134182035923004) (24, 0.41087019443511963)
      };
      \addlegendentry{\scalebox{.7}{\textsc{Eit}-24L}}

      \addplot[thick, blue, mark=none, mark size=1pt, thick, mark options={solid}] coordinates {
(1, 0.06176108866930008) (2, 0.08197025209665298) (3, 0.10155477374792099) (4, 0.122928187251091) (5, 0.13654619455337524) (6, 0.1483166217803955) (7, 0.14987294375896454) (8, 0.15625575184822083) (9, 0.1636214256286621) (10, 0.16641561686992645) (11, 0.16659535467624664) (12, 0.17167291045188904) (13, 0.17107966542243958) (14, 0.1715303212404251) (15, 0.17578622698783875) (16, 0.17731110751628876) (17, 0.1883384734392166) (18, 0.18920697271823883) (19, 0.2000872790813446) (20, 0.20473994314670563) (21, 0.21846477687358856) (22, 0.2235085368156433) (23, 0.22809936106204987) (24, 0.2335023581981659) (25, 0.24021390080451965) (26, 0.24500620365142822) (27, 0.25562360882759094) (28, 0.26059794425964355) (29, 0.26554539799690247) (30, 0.26366928219795227) (31, 0.26836469769477844) (32, 0.27560192346572876) (33, 0.2783666253089905) (34, 0.2782844007015228) (35, 0.2787834107875824) (36, 0.2830398380756378) (37, 0.2859547436237335) (38, 0.2896917462348938) (39, 0.29959943890571594) (40, 0.30580341815948486) (41, 0.317562073469162) (42, 0.32888683676719666) (43, 0.3428316116333008) (44, 0.3590424060821533) (45, 0.37961772084236145) (46, 0.39686423540115356) (47, 0.4127797484397888) (48, 0.4252675771713257)
};
\addlegendentry{\scalebox{.7}{\textsc{Eit}-48L}}

 \end{axis}
}
\end{tikzpicture}

\vskip -0.1in
\caption{Analysis of token correlation on En-De task.}
\label{fig:token_correlation_measurement}
\end{figure}

%% file: Figure/Analysis/efficient_comparision.tex
\definecolor{tiffanyblue}{RGB}{129,216,208}
\definecolor{bangdiblue}{RGB}{0,149,182}
\definecolor{kleinblue}{RGB}{0,47,167}
\definecolor{kabuliblue}{RGB}{26,85,153}
\definecolor{purple}{RGB}{138,43,226}
\usetikzlibrary{matrix}
\begin{figure*}[ht]
  \centering

  \begin{tikzpicture}[]
  \pgfplotsset{set layers}

      \scriptsize{
      \begin{axis}[
	 at={(0,0)},
      ymajorgrids,
      xmajorgrids,
      grid style=dashed,
      width=.5\textwidth,
      height=.28\textwidth,
      legend style={at={(0.23,0.08)}, anchor=south west},
      xlabel={\small{Encoder Depth}},
      ylabel={\small{Mem.(G)}},
      ylabel style={yshift=-2em},xlabel style={yshift=1.0em},
      yticklabel style={/pgf/number format/precision=1,/pgf/number format/fixed zerofill},
      ymin=10,ymax=19.5, ytick={10, 13,16,19},
      xmin=3,xmax=27,xtick={ 6, 12, 18, 24},
      legend style={draw=none,fill=none,yshift=31pt,xshift=-5.1em, legend plot pos=right,font={\footnotesize},cells={anchor=west}, legend columns={1}}
      ]
      \addplot[blue,mark=diamond*,mark size=1.5pt,thick,mark options={fill=blue,draw=blue,line width=0.125pt}] coordinates { (6,10.60) (12,11.45) (18,12.41) (24,14.01)
      };
      \addlegendentry{\scalebox{1}{Transformer}}

      \addplot[teal!70,mark=pentagon*,mark size=1.5pt,thick,mark options={fill=teal,draw=teal,line width=0.125pt}] coordinates {(6,11.10)  (12,12.55) (18,13.95) (24,15.53)};
    \addlegendentry{\scalebox{1}{\textsc{E-Eit}}}
      
      \addplot[red!60,mark=*,mark size=1.7pt,thick,mark options={fill=red,draw=red,line width=0.125pt}] coordinates {(6,11.50)  (12,14.50) (18,15.70) (24,18.16)};
      \addlegendentry{\scalebox{1}{\textsc{Eit}}}

      \end{axis}

     }

     \scriptsize{
      \begin{axis}[
	 at={(0.5*\textwidth,0)},
      ymajorgrids,
      xmajorgrids,
      grid style=dashed,
      width=.5\textwidth,
      height=.28\textwidth,
      legend style={at={(0.23,0.08)}, anchor=south west},
      xlabel={\small{Encoder Depth}},
      ylabel={\small{Epoch (s)}},
      ylabel style={yshift=-1.5em},xlabel style={yshift=1.0em},
      yticklabel style={/pgf/number format/precision=1,/pgf/number format/fixed zerofill},
      ymin=9,ymax=170, ytick={10, 60, 110, 160},
      xmin=3,xmax=51,xtick={ 6, 24, 36, 48},
      legend style={draw=none,fill=none,yshift=31pt,xshift=-5.1em, legend plot pos=right,font={\footnotesize},cells={anchor=west}, legend columns={1}}
      ]

      \addplot[blue,mark=diamond*,mark size=1.5pt,thick,mark options={fill=blue,draw=blue,line width=0.125pt}] coordinates {(6,20)  (24,40) (36,50) (48, 80)
      };\label{Baseline}
      \addlegendentry{\scalebox{1}{Transformer}}

      \addplot[teal!70,mark=pentagon*,mark size=1.5pt,thick,mark options={fill=teal,draw=teal,line width=0.125pt}] coordinates {(6,22)  (24,50) (36,64) (48, 99)};\label{E-Eit}
      \addlegendentry{\scalebox{1}{\textsc{E-Eit}}}

          \addplot[red!60,mark=*,mark size=1.7pt,thick,mark options={fill=red,draw=red,line width=0.125pt}] coordinates {(6,28)  (24,78) (36,113) (48, 157)};\label{Eit}
      \addlegendentry{\scalebox{1}{\textsc{Eit}}}

      \end{axis}

     }

  \end{tikzpicture}
  \vskip -0.1in
  \caption{Memory and speed vs. encoder depth. \textsc{E-Eit} can achieve comparable results with fewer training costs than \textsc{Eit}. }\label{fig:efficient_compariosion_en_de}
\end{figure*}

%% file: Figure/Analysis/visual_ppl.tex
\definecolor{tiffanyblue}{RGB}{129,216,208}
\definecolor{bangdiblue}{RGB}{0,149,182}
\definecolor{kleinblue}{RGB}{0,47,167}
\definecolor{kabuliblue}{RGB}{26,85,153}
\definecolor{purple}{RGB}{138,43,226}
\usetikzlibrary{matrix}
\definecolor{Magenta2}{RGB}{238, 0, 238}
\definecolor{ugreen}{RGB}{118,218,145}
\definecolor{darkgreen}{RGB}{179, 214,110}
\definecolor{uyellow}{RGB}{239,202,102}
\definecolor{ublue}{RGB}{207,234,241}
\definecolor{upurple}{RGB}{196,165,222}
\definecolor{uorange}{RGB}{248,172,140}
\definecolor{upink}{RGB}{255,150,128}
\usetikzlibrary {patterns,patterns.meta}
\definecolor{palewhite}{RGB}{137, 157, 192}

\usetikzlibrary {patterns,patterns.meta}
\begin{figure*}[ht]

  \centering

  \begin{tikzpicture}[]
  \pgfplotsset{set layers}
    \scriptsize{
      \begin{axis}[
	 at={(0.5\textwidth,0)},
      ymajorgrids,
      xmajorgrids,
      grid style=dashed,
      width=.5\textwidth,
      height=.28\textwidth,
      legend style={at={(0.23,0.08)}, anchor=south west},
      xlabel={\scriptsize{Epoch}},
      ylabel={\scriptsize{Validation Perplexity}},
      ylabel style={yshift=-2em},xlabel style={yshift=1.0em},
      yticklabel style={/pgf/number format/precision=1,/pgf/number format/fixed zerofill},
      ymin=3.5,ymax=4.5, ytick={3.5,4,4.5},
      xmin=5,xmax=22,xtick={5, 10, 15, 20 },
      legend style={draw=none,fill=none,yshift=3.5em,xshift=8.6em, inner sep=1pt, legend plot pos=right,cells={anchor=west}}
      ]

    \addplot[blue,mark=*,mark size=1pt,thick,mark options={fill=blue,draw=blue,line width=1.25pt}] coordinates { (1,17.81) (2,6.83) (3,5.58) (4,5.08) (5,4.83) (6, 4.64) (7,4.43) (8,4.29) (9,4.19) (10,4.08) (11,3.99) (12,3.94) (13,3.88) (14,3.88) (15,3.83) (16,3.81) (17, 3.77) (18,3.77) (19,3.75) (20,3.71) (21,3.73)
      };
      \addlegendentry{\scalebox{1}{Transformer}}

      \addplot[red,mark=diamond*,mark size=1pt,thick,mark options={fill=red,draw=red,line width=1.25pt}] coordinates { (1,17.3) (2,6.8) (3,5.41) (4,4.97) (5,4.71) (6, 4.49) (7,4.31) (8,4.18) (9,4.1) (10,4) (11,3.92) (12,3.87) (13,3.82) (14,3.83) (15,3.78) (16,3.77) (17,3.71) (18,3.7) (19,3.69) (20,3.65) (21,3.68)
      };
      \addlegendentry{\scalebox{1}{\textsc{Eit}}}
      
      \end{axis}

     }

      \scriptsize{
      \begin{axis}[
	 at={(0,0)},
      ymajorgrids,
      xmajorgrids,
      grid style=dashed,
      width=.5\textwidth,
      height=.28\textwidth,
      legend style={at={(0.23,0.08)}, anchor=south west},
      xlabel={\scriptsize{Epoch}},
      ylabel={\scriptsize{Training Perplexity}},
      ylabel style={yshift=-2em},xlabel style={yshift=1.0em},
      yticklabel style={/pgf/number format/precision=1,/pgf/number format/fixed zerofill},
      ymin=6,ymax=9, ytick={6, 7, 8,9},
      xmin=5,xmax=22,xtick={5, 10, 15, 20 },
      legend style={draw=none,fill=none,yshift=3.5em,xshift=8.6em, inner sep=1pt, legend plot pos=right,cells={anchor=west}}
      ]

     \addplot[blue,mark=*,mark size=1pt,thick,mark options={fill=blue,draw=blue,line width=1.25pt}] coordinates { (1,297.89) (2,17.79) (3,11.64) (4,10.05) (5,9.31) (6, 8.87) (7,8.55) (8,8.15) (9,7.81) (10,7.56) (11,7.37) (12,7.22) (13,7.1) (14,7) (15,6.91) (16,6.84) (17,6.78) (18,6.72) (19,6.67) (20,6.63) (21, 6.59)
      };
      \addlegendentry{\scalebox{1}{Transformer}}

      \addplot[red,mark=diamond*,mark size=1pt,thick,mark options={fill=red,draw=red,line width=1.25pt}] coordinates { (1,276.1) (2,17.65) (3,11.46) (4,9.84) (5,9.09) (6, 8.65) (7,8.32) (8,7.93) (9,7.6) (10,7.36) (11,7.18) (12,7.04) (13,6.93) (14,6.83) (15,6.75) (16,6.68) (17,6.62) (18,6.56) (19,6.52) (20,6.47) (21,6.43)
      };
      \addlegendentry{\scalebox{1}{\textsc{Eit}}}

      \end{axis}

     }

  \end{tikzpicture}

  \vspace{-1em}
  \caption{Training perplexity and validation perplexity of Transformer and our \textsc{Eit} on WMT'14 En-De task. Note that the models are in \textit{base} configuration. }\label{fig:visual_ppl}
\end{figure*}
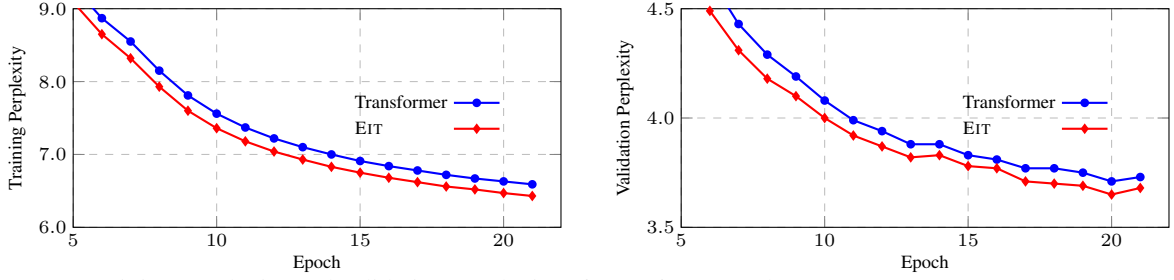

%% file: Figure/Analysis/parameter_analysis.tex
\definecolor{tiffanyblue}{RGB}{129,216,208}
\definecolor{bangdiblue}{RGB}{0,149,182}
\definecolor{kleinblue}{RGB}{0,47,167}
\definecolor{kabuliblue}{RGB}{26,85,153}
\definecolor{purple}{RGB}{138,43,226}
\usetikzlibrary{matrix}
\begin{figure*}[ht]
  \centering

  \begin{tikzpicture}[]
  \pgfplotsset{set layers}
    \node[font=\footnotesize] at (3.2em, -2.3em-0.14*\textwidth){(a)};
    \node[font=\footnotesize] at (13.2em, -2.3em-0.14*\textwidth){(b)};
    \node[font=\footnotesize] at (23.2em, -2.3em-0.14*\textwidth){(c)};
    \node[font=\footnotesize] at (33.2em, -2.3em-0.14*\textwidth){(d)};

	\scriptsize{
      \begin{axis}[
	 at={(0,-.14\textwidth)},
      ymajorgrids,
      xmajorgrids,
      grid style=dashed,
      width=.28\textwidth,
      height=.23\textwidth,
      legend style={at={(0.23,0.08)}, anchor=south west},
      xlabel={\scriptsize{ISI Kernel Size}},
      ylabel={\scriptsize{BLEU}},
      ylabel style={yshift=-2em},xlabel style={yshift=1.0em},
      yticklabel style={/pgf/number format/precision=1,/pgf/number format/fixed zerofill},
      ymin=26.8,ymax=28.2, ytick={27,27.5,28},
      xmin=0,xmax=10,xtick={1,3, 5, 7, 9},
      legend style={draw=none,fill=none,yshift=-6pt,xshift=-2em, legend plot pos=right,font={\footnotesize},cells={anchor=west}}
      ]
      \addplot[blue,dashed, thick,mark options={fill=white,draw=red,line width=1.0pt}] coordinates { (0.5,27.13)  (2,27.13) (4,27.13) (9.7,27.13)
      };

      \addplot[red!60,mark=*,mark size=1.7pt,thick,mark options={fill=red,draw=red,line width=0.5pt}] coordinates {(1,27.37)  (3,27.61) (5,27.86) (7,27.96) (9,27.71)};
      
      \end{axis}

     }
     	\scriptsize{
      \begin{axis}[
	 at={(.25\textwidth,-.14\textwidth)},
      ymajorgrids,
      xmajorgrids,
      grid style=dashed,
      width=.28\textwidth,
      height=.23\textwidth,
      legend style={at={(0.23,0.08)}, anchor=south west},
      xlabel={\scriptsize{CSI Kernel Size}},
      ylabel={\scriptsize{BLEU}},
      ylabel style={yshift=-2em},xlabel style={yshift=1.0em},
      yticklabel style={/pgf/number format/precision=1,/pgf/number format/fixed zerofill},
      ymin=26.8,ymax=28.2, ytick={27,27.5,28},
      xmin=0,xmax=10,xtick={1,3, 5, 7, 9},
      legend style={draw=none,fill=none,yshift=8.5em,xshift=2em, legend plot pos=right,font={\footnotesize},cells={anchor=west}},
       legend columns=2
      ]
      \addplot[blue,dashed, thick,mark options={fill=white,draw=red,line width=1.0pt}] coordinates { (0.5,27.13)  (2,27.13) (4,27.13) (9.5,27.13)
      };
\addlegendentry{\scalebox{.8}{Transformer}}

    \addplot[red!60,mark=*,mark size=1.7pt,thick,mark options={fill=red,draw=red,line width=0.5pt}] coordinates {(1,27.96)  (3,28.00) (5,27.66) (7,27.73) (9,27.72)};
    \addlegendentry{\scalebox{.8}{\textsc{Eit}}}

      \end{axis}

     }
     	\scriptsize{
      \begin{axis}[
	 at={(.5\textwidth,-.14\textwidth)},
      ymajorgrids,
      xmajorgrids,
      grid style=dashed,
      width=.28\textwidth,
      height=.23\textwidth,
      legend style={at={(0.23,0.08)}, anchor=south west},
      xlabel={\scriptsize{$M^{H_{isi}}/M^2$}},
      ylabel={\scriptsize{BLEU}},
      ylabel style={yshift=-2em},xlabel style={yshift=1.3em},
      yticklabel style={/pgf/number format/precision=1,/pgf/number format/fixed zerofill},
      ymin=26.8,ymax=28.2, ytick={27,27.5,28},
      xmin=0.5,xmax=4.5,xtick={1,2,4},
      legend style={yshift=2em,xshift=-2em, legend plot pos=right,font={\footnotesize},cells={anchor=west}}
      ]
      \addplot[blue,dashed, thick,mark options={fill=white,draw=blue,line width=1.0pt}] coordinates { (0.75,27.13)  (2,27.13) (4,27.13) (4.25,27.13)
      };

     \addplot[red!60,mark=*,mark size=1.7pt,thick,mark options={fill=red,draw=red,line width=0.5pt}] coordinates {(1,27.81)  (2,28.00) (4,27.66)};
      \end{axis}

     }
     	\scriptsize{
      \begin{axis}[
	 at={(.75\textwidth,-.14\textwidth)},
      ymajorgrids,
      xmajorgrids,
      grid style=dashed,
      width=.28\textwidth,
      height=.23\textwidth,
      legend style={at={(0.23,0.08)}, anchor=south west},
      xlabel={\scriptsize{$M^{H_{csi}}/M$}},
      ylabel={\scriptsize{BLEU}},
      ylabel style={yshift=-2em},xlabel style={yshift=1.3em},
      yticklabel style={/pgf/number format/precision=1,/pgf/number format/fixed zerofill},
      ymin=26.8,ymax=28.2, ytick={27,27.5,28},
      xmin=0,xmax=18,xtick={1, 4, 8, 16},
      legend style={yshift=-6pt,xshift=-2em, legend plot pos=right,font={\footnotesize},cells={anchor=west}}
      ]
      \addplot[blue,dashed, thick,mark options={fill=white,draw=blue,line width=1.0pt}] coordinates {(0.5, 27.13)(1,27.13) (2, 27.13)(3,27.13) (4,27.13) (8.5,27.13) (17,27.13)
      };

      \addplot[red!60,mark=*,mark size=1.7pt,thick,mark options={fill=red,draw=red,line width=0.5pt}] coordinates {(1, 27.52 ) (4,27.66)  (8,28.00) (16,27.81)};

      \end{axis}

     }
     
  \end{tikzpicture}
  \vskip -0.1in
  \caption{The comparison of BLEU against different hyper-parameters. Note that the blue horizontal line represents the performance of Transformer.}\label{fig:parameter_analysis}
\end{figure*}
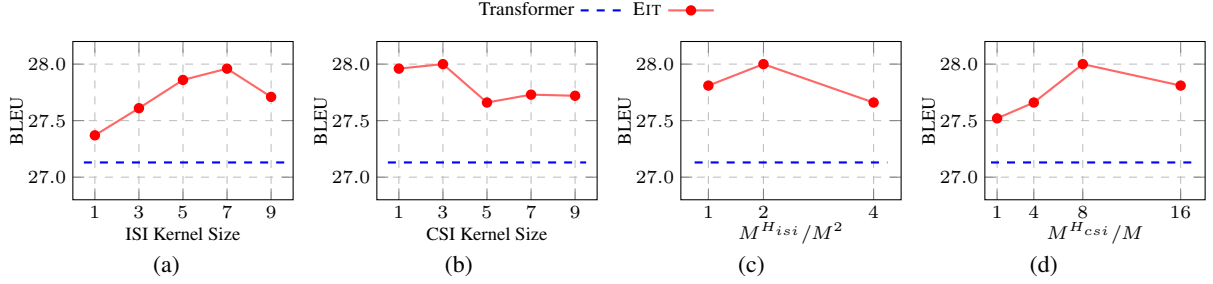

%% file: Figure/Analysis/local_analysis.tex
\definecolor{tiffanyblue}{RGB}{129,216,208}
\definecolor{bangdiblue}{RGB}{0,149,182}
\definecolor{kleinblue}{RGB}{0,47,167}
\definecolor{kabuliblue}{RGB}{26,85,153}
\definecolor{purple}{RGB}{138,43,226}
\usetikzlibrary{matrix}
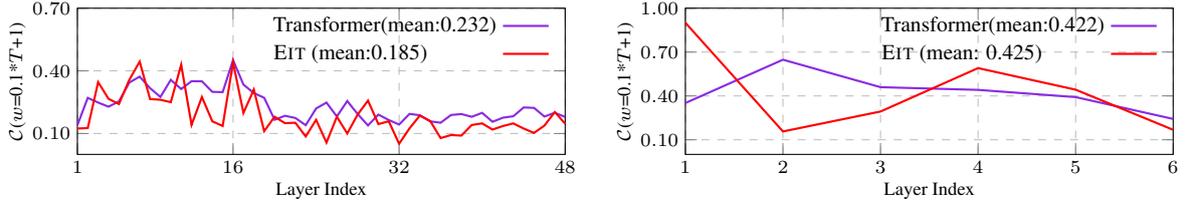
\begin{figure*}[t!]
\vspace{-0.1in}
  \centering
  \tikzset{global scale/.style={
    scale=#1,
    every node/.append style={scale=#1}
  }
}
  \begin{tikzpicture}[global scale=1]
  \pgfplotsset{set layers}

     	\scriptsize{
      \begin{axis}[
	 at={(0,0)},
      ymajorgrids,
      xmajorgrids,
      grid style=dashed,
      width=0.5\textwidth,
      height=.22\textwidth,
      legend style={at={(0.2,0.05)}, anchor=south west},
      xlabel={\scriptsize{Layer Index}},
      ylabel={\scriptsize{$\mathcal{C}$($w$=0.1*$T$+1)}},
      ylabel style={yshift=-1.8em},xlabel style={yshift=1.0em},
      yticklabel style={/pgf/number format/precision=2,/pgf/number format/fixed zerofill},
      ymin=0,ymax=0.7, ytick={0.1,0.4, 0.7},
      xmin=1,xmax=48,xtick={1,16, 32, 48},
      legend style={draw=none,fill=none, yshift=4.1em,xshift=4.8em,inner sep=1pt,legend plot pos=right,font={\footnotesize},cells={anchor=west}}
      ]

      \addplot[purple,thick] coordinates { (1,0.13974395394325256)
(2,0.2707097828388214)
(3,0.24827298521995544)
(4,0.22841940820217133)
(5,0.25355949997901917)
(6,0.3439560532569885)
(7,0.3725948929786682)
(8,0.3172502815723419)
(9,0.27509868144989014)
(10,0.35708504915237427)
(11,0.3127879202365875)
(12,0.3505582809448242)
(13,0.3499607443809509)
(14,0.29945170879364014)
(15,0.29814574122428894)
(16,0.45164015889167786)
(17,0.3342216908931732)
(18,0.2938527762889862)
(19,0.27102237939834595)
(20,0.16567619144916534)
(21,0.18499404191970825)
(22,0.1744394302368164)
(23,0.1397840827703476)
(24,0.21991311013698578)
(25,0.24815362691879272)
(26,0.18153312802314758)
(27,0.25553634762763977)
(28,0.19367410242557526)
(29,0.13853059709072113)
(30,0.1907797008752823)
(31,0.1647561937570572)
(32,0.14299693703651428)
(33,0.1939365565776825)
(34,0.1874580681324005)
(35,0.15810592472553253)
(36,0.15154923498630524)
(37,0.18858936429023743)
(38,0.1931883990764618)
(39,0.1801261603832245)
(40,0.19889387488365173)
(41,0.156203031539917)
(42,0.1758049577474594)
(43,0.1826404184103012)
(44,0.2248682826757431)
(45,0.2226613610982895)
(46,0.18123574554920197)
(47,0.19930806756019592)
(48,0.1794007122516632)
      };\label{baseline_48_attn_localness_w_0.1}
      \addlegendentry{\scalebox{.9}{Transformer(mean:0.232)}}

      \addplot[red, thick] coordinates { (1,0.12398918718099594)
(2,0.12606042623519897)
(3,0.3459605872631073)
(4,0.26601001620292664)
(5,0.24105115234851837)
(6,0.35693588852882385)
(7,0.4441947937011719)
(8,0.26522254943847656)
(9,0.2616423964500427)
(10,0.24991492927074432)
(11,0.42946693301200867)
(12,0.1414141058921814)
(13,0.27475038170814514)
(14,0.15843233466148376)
(15,0.13593485951423645)
(16,0.44041016697883606)
(17,0.1982985883951187)
(18,0.31020253896713257)
(19,0.1111919954419136)
(20,0.18033351004123688)
(21,0.1495506465435028)
(22,0.15165939927101135)
(23,0.08583426475524902)
(24,0.16546407341957092)
(25,0.05653684586286545)
(26,0.18022006750106812)
(27,0.100051648914814)
(28,0.1906697154045105)
(29,0.2575818598270416)
(30,0.14501921832561493)
(31,0.15874023735523224)
(32,0.05095166712999344)
(33,0.12456435710191727)
(34,0.18586429953575134)
(35,0.15967530012130737)
(36,0.07811659574508667)
(37,0.0926671177148819)
(38,0.09005400538444519)
(39,0.14089182019233704)
(40,0.14844810962677002)
(41,0.11884474754333496)
(42,0.1358156055212021)
(43,0.14841291308403015)
(44,0.1237628236413002)
(45,0.1028289645910263)
(46,0.13803769648075104)
(47,0.20302341878414154)
(48,0.1467488557100296)
      };\label{eit_48_attn_localness_w_0.1}
      \addlegendentry{\scalebox{.9}{\textsc{Eit} (mean:0.185)}}

      \end{axis}

     }

       \scriptsize{
      \begin{axis}[
	 at={(0.5\textwidth,0)},
      ymajorgrids,
      xmajorgrids,
      grid style=dashed,
      width=0.5\textwidth,
      height=.22\textwidth,
      legend style={at={(0.2,0.05)}, anchor=south west},
      xlabel={\scriptsize{Layer Index}},
      ylabel={\scriptsize{$\mathcal{C}$($w$=0.1*$T$+1)}},
      ylabel style={yshift=-1.8em},xlabel style={yshift=1.0em},
      yticklabel style={/pgf/number format/precision=2,/pgf/number format/fixed zerofill},
      ymin=0,ymax=1, ytick={0.1,0.4, 0.7, 1.0},
      xmin=1,xmax=6,xtick={1,2, 3, 4, 5, 6},
      legend style={draw=none,fill=none, yshift=4.1em,xshift=4.8em,inner sep=1pt,legend plot pos=right,font={\footnotesize},cells={anchor=west}}
      ]

                 \addplot[purple,thick] coordinates { 
(1,0.35139647126197815)
(2,0.6478139162063599)
(3,0.45979756116867065)
(4,0.4412798285484314)
(5,0.39195311069488525)
(6,0.24244631826877594)

      };\label{baseline-6}
   \addlegendentry{\scalebox{.9}{Transformer(mean:0.422)}}
       \addplot[red, thick] coordinates { 
(1,0.9005311727523804)
(2,0.1569214016199112)
(3,0.2933032512664795)
(4,0.5902840495109558)
(5,0.4427495002746582)
(6,0.1680721640586853)

      };\label{eit-6}
       \addlegendentry{\scalebox{.9}{\textsc{Eit} (mean: 0.425)}}

      \end{axis}

     }

  \end{tikzpicture}

    \vskip -0.1in
  \caption{Quantitative analysis on localness in attention maps on En-De task (Above) and CNN-DailyMail task (Bottom).}\label{fig:localness_measure}
\end{figure*}

%% file: Figure/Analysis/dynamics_similarity.tex
\definecolor{tiffanyblue}{RGB}{129,216,208}
\definecolor{bangdiblue}{RGB}{0,149,182}
\definecolor{kleinblue}{RGB}{0,47,167}
\definecolor{kabuliblue}{RGB}{26,85,153}
\definecolor{purple}{RGB}{138,43,226}
\usetikzlibrary{matrix}
\begin{figure}[ht]
  \centering
  \begin{tikzpicture}[]
  \pgfplotsset{set layers}

     	\scriptsize{
      \begin{axis}[
	 at={(0,0)},
      ymajorgrids,
      xmajorgrids,
      grid style=dashed,
      width=0.5\textwidth,
      height=.23\textwidth,
      legend style={at={(0.23,0.08)}, anchor=south west},
      xlabel={\small{Layer Index}},
      ylabel={\small{Head Similarity}},
      ylabel style={yshift=-1.8em},xlabel style={yshift=1.0em},
      yticklabel style={/pgf/number format/precision=2,/pgf/number format/fixed zerofill},
      ymin=0,ymax=1, ytick={0.1,0.4, 0.7, 1.0},
      xmin=1,xmax=48,xtick={1,16, 32, 48},
      legend style={yshift=-6pt,xshift=-2em, legend plot pos=right,font={\small},cells={anchor=west}}
      ]

      \addplot[thick,blue] coordinates { 
(1,0.7266814112663269)
(2,0.6880645751953125)
(3,0.5206577181816101)
(4,0.4719516336917877)
(5,0.5762128829956055)
(6,0.5601311922073364)
(7,0.4896090030670166)
(8,0.47644278407096863)
(9,0.2894037365913391)
(10,0.5933544039726257)
(11,0.44920966029167175)
(12,0.3270474076271057)
(13,0.3398246467113495)
(14,0.16641971468925476)
(15,0.19950354099273682)
(16,0.2153274267911911)
(17,0.16155129671096802)
(18,0.17437447607517242)
(19,0.10965654253959656)
(20,0.10872341692447662)
(21,0.09126676619052887)
(22,0.15945814549922943)
(23,0.10356763005256653)
(24,0.12222781777381897)
(25,0.10653897374868393)
(26,0.11253912001848221)
(27,0.09545107185840607)
(28,0.08702044934034348)
(29,0.10466318577528)
(30,0.09330723434686661)
(31,0.09158007055521011)
(32,0.07538644224405289)
(33,0.08136588335037231)
(34,0.11582925915718079)
(35,0.06886392831802368)
(36,0.08172079175710678)
(37,0.06505969166755676)
(38,0.06882491707801819)
(39,0.08823312819004059)
(40,0.06944296509027481)
(41,0.06318416446447372)
(42,0.1003325879573822)
(43,0.07473566383123398)
(44,0.07320964336395264)
(45,0.06801886856555939)
(46,0.08881758153438568)
(47,0.08708243817090988)
(48,0.07032446563243866)

      };\label{after_rfe_48_attn_distance}

\node[anchor=west,font=\tiny] at (axis cs:32, 0.25) {\bf After M2M };
    \draw[<-, thick] (axis cs:27,0.09545107185840607) -- (axis cs:32, 0.25) ;

      \addplot[thick,tiffanyblue] coordinates { 
      
      (1,0.43063732981681824)
(2,0.3699701130390167)
(3,0.17853717505931854)
(4,0.14477993547916412)
(5,0.25572657585144043)
(6,0.15324446558952332)
(7,0.15243802964687347)
(8,0.15223592519760132)
(9,0.12721334397792816)
(10,0.2444162666797638)
(11,0.24008092284202576)
(12,0.17474865913391113)
(13,0.11723438650369644)
(14,0.09759797155857086)
(15,0.11657334119081497)
(16,0.1419530063867569)
(17,0.1450585275888443)
(18,0.1305314153432846)
(19,0.08186163008213043)
(20,0.09745525568723679)
(21,0.10220438241958618)
(22,0.11726595461368561)
(23,0.1285841166973114)
(24,0.07868873327970505)
(25,0.09603174775838852)
(26,0.0876379981637001)
(27,0.08229043334722519)
(28,0.11775492876768112)
(29,0.10171419382095337)
(30,0.13264551758766174)
(31,0.09796257317066193)
(32,0.10576394945383072)
(33,0.18759173154830933)
(34,0.07880754768848419)
(35,0.09168055653572083)
(36,0.0673186406493187)
(37,0.0669429823756218)
(38,0.07506689429283142)
(39,0.07825274020433426)
(40,0.06509970128536224)
(41,0.08466073125600815)
(42,0.0892268642783165)
(43,0.07363972067832947)
(44,0.10268931090831757)
(45,0.07526622712612152)
(46,0.12223149091005325)
(47,0.060859743505716324)
(48,0.06454169750213623)

      };\label{after_isi_48_attn_distance}

      \node[anchor=west,font=\tiny] at (axis cs:16, 0.3) {\bf After ISI };

      \draw[<-, thick] (axis cs:11,0.24008092284202576) -- (axis cs:16, 0.3) ;

      \addplot[thick,red] coordinates { (1, 0.8723) (2,0.7969) (3,0.7914) (4,0.7271) (5,0.6798) (6,0.5685) (7,0.6669) (8,0.7418)(9,0.7173)(10,0.3784)
      (11,0.6357)(12,0.7789) (13,0.4224) (14,0.6907) (15,0.5423) (16,0.5062) (17,0.4817) (18,0.7720) (19,0.5264) (20,0.5562) (21,0.7418) (22,0.5906) (23,0.7000) (24,0.5888) (25,0.7439) (26,0.6332) (27,0.6911) (28,0.6641) (29,0.5506) (30,0.5958) (31,0.5898) (32,0.7075) (33,0.7032) (34,0.5175) (35,0.7269) (36,0.6401) (37,0.8256) (38,0.8022) (39,0.6870) (40,0.7196) (41,0.8104) (42,0.4825) (43,0.6597) (44,0.7039) (45,0.7645) (46,0.5173) (47,0.5548) (48,0.6705)
      };\label{eit_48_attn_distance}

            \node[anchor=west,font=\tiny] at (axis cs:20, 0.9) {\bf After CSI };
                  
    \draw[<-, thick] (axis cs:12,0.7789) -- (axis cs:20, 0.9) ;

      \end{axis}

     }

  \end{tikzpicture}

  \vskip -0.1in
  \caption{Dynamics of attention map similarity.}\label{fig:dynamics_similarity}
\end{figure}
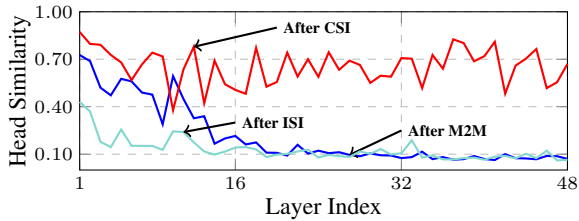